\documentclass[twocolumn]{svjour3}
\smartqed

\usepackage{float}
\usepackage{graphicx}
\usepackage{natbib}
\usepackage{setspace}
\usepackage{color}
\usepackage{graphbox}

\usepackage{url}
\usepackage{amsmath,amssymb}

\usepackage{ntheorem}
\usepackage{color}
\usepackage[colorlinks=true,linkcolor=black]{hyperref}
\usepackage{breakurl}
\usepackage[ruled,linesnumbered]{algorithm2e}
\usepackage{multicol}
\usepackage{lipsum}
\usepackage{algorithmic}

\usepackage{subcaption}
\usepackage{epstopdf}
\usepackage{enumerate}
\usepackage{enumitem}

\usepackage{amsmath}

\DeclareMathOperator*{\argmax}{arg\,max}
\newcommand{\supemph}[1]{\textbf{\emph {#1}}}

\begin{document}

\title{Vision-Based Preharvest Yield Mapping for Apple Orchards}

\author{Pravakar Roy \and Abhijeet Kislay \and Patrick A. Plonski \and James Luby \and Volkan Isler
}


\institute{ Pravakar Roy \and Abhijeet Kislay \and Patrick A. Plonski \and Vokan Isler \at
              Robotic Sensor Networks (RSN) Lab, Department of Computer Science and Engineering, University of Minnesota \\
              \email{\{royxx268, kislay004, plons005, isler\}@umn.edu}           
           \and
           James Luby \at Department of Horticultural Science, University of Minnesota\\
           \email{lubyx001@umn.edu}
           }

\maketitle
\begin{abstract}
We present an end-to-end computer vision system for mapping yield in an apple orchard using images captured from a single camera. Our proposed system is platform independent and does not require any specific lighting conditions. Our main technical contributions are 1)~a semi-supervised clustering algorithm that utilizes colors to identify apples and 2)~an unsupervised clustering method that utilizes spatial properties to estimate fruit counts from apple clusters having arbitrarily complex geometry. Additionally, we utilize camera motion to merge the counts across multiple views. We verified the performance of our algorithms by conducting multiple field trials on three tree rows consisting of $252$ trees at the University of Minnesota Horticultural Research Center. Results indicate that the detection method achieves $F_1$-measure $.95 -.97$ for multiple color varieties and lighting conditions. The counting method achieves an accuracy of $89\%-98\%$. Additionally, we report merged fruit counts from both sides of the tree rows. Our yield estimation method achieves an overall accuracy of $91.98\% - 94.81\%$ across different datasets.

\end{abstract}

\bibliographystyle{spbasic}

Efficient control and management of agricultural farms are essential to cope with growing population and numerous environmental and economic issues. Precision agriculture techniques provide farmers with methods to monitor the status of their crops on demand and enable them to make crucial decisions (trimming, pruning, application of fertilizers and pesticides etc.). For many commodity crops (rice, corn, wheat etc.), these techniques are already mature enough to provide on-demand maps (yield, crop stress, water quality etc.) as well as help create an infrastructure to implement crucial tasks. However, for specialty crops (such as fruits, vegetables, and flowers), these techniques are still evolving. Specialty crops are particularly good candidates for precision farming and phenotyping studies, because of their species diversity, high value, high management cost, and high variability in growth.  One of the main bottlenecks for both these applications is the lack of a convenient yield mapping system.

The variability in plant size, color, shape etc., make it difficult to develop a general yield mapping method. Instead, researchers developed yield mapping systems specific to each crop (\cite{senthilnath2016detection,wang,das2015devices,gongal2016apple,silwal2014apple}). In this paper, we focus on apple orchards and present a complete system for yield monitoring. Our system has  a couple of distinguishing features:
\begin{enumerate}
\item It is platform independent and relies only on images captured from a single camera. The camera can be mounted on a ground robot or an Unmanned Aerial Vehicle (UAV). It can also be handheld.
\item It does not require any special lighting condition. It operates at daytime in a completely natural environment that is convenient for field applications.
\end{enumerate}

\begin{figure}[!hbpt]
\includegraphics[width = .18\columnwidth]{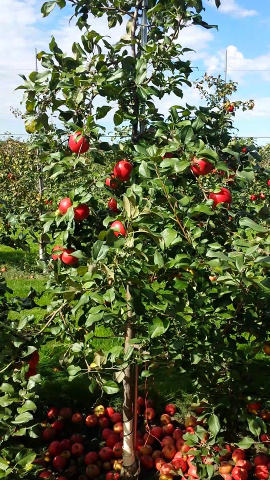}          
\includegraphics[width = .18\columnwidth]{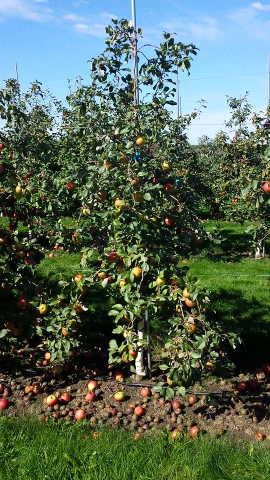}           
\includegraphics[width = .18\columnwidth]{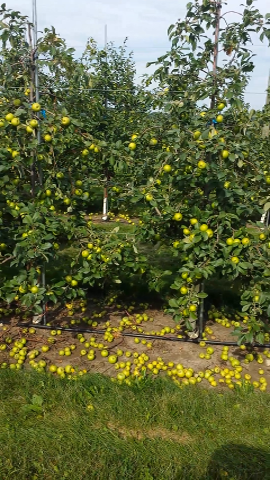}           
\includegraphics[width = .18\columnwidth]{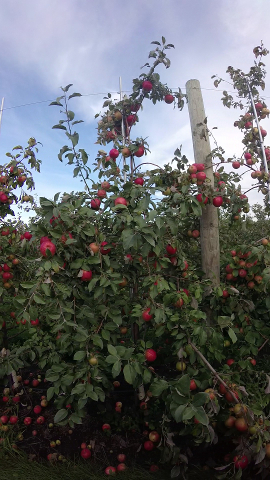}           
\includegraphics[width = .18\columnwidth]{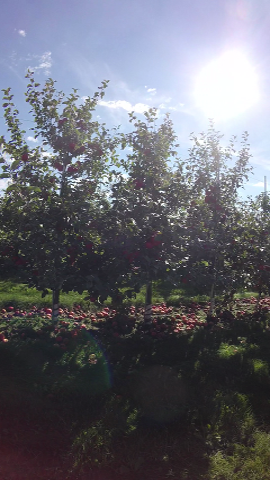}           
   \caption{Conditions typical to orchard settings. Our algorithms perform well in the first four cases. For the final image, it requires a significant amount of user supervision to obtain reasonable performance.}
   \label{fig:Conditions}
\end{figure} 

\begin{figure}[!hbpt]
{\centering
\includegraphics[height=2cm]{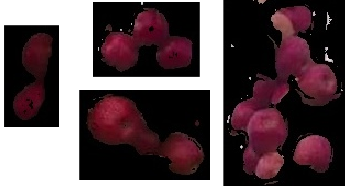}
\vrule 
\includegraphics[height=2cm]{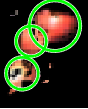}

\caption{
From left to right: Segmented images exhibiting complex geometry. Sample output of our algorithm on an individual cluster. \label{fig:counting_ex}}
}
\end{figure}

We have two main technical contributions in this paper. 1) We present a novel semi-supervised segmentation method to separate the apple pixels from others in the input images. Varying colors of apples (red, golden, green, yellow along with the mixture of their shades), different lighting conditions (based on the position of the sun, clouds etc.), shadows and occlusions created by leaves and branches make the job of identifying regions belonging to apples in an image difficult. We present a semi-supervised segmentation method (Fig.~\ref{fig:seg_pipeline}) which utilizes minimal user interaction to train a classification model and uses it to identify the apples. Additionally, the system is capable of storing trained models that can be used for a similar variety of apples and lighting conditions. A key aspect of our segmentation method is that it does not rely on detecting every apple in a single frame (Fig.~\ref{fig:eval_metric}(\subref{fig:inherent_fuziness})). We utilize multiple views for detection and counting based on the assumption that for every apple there is ``some'' view from which it can be detected. This provides robustness against false positives (caused by shadows and specularities) and helps us achieve high precision and recall (Fig.~\ref{fig:prec},\ref{fig:rec}). We investigate the sensitivity of our detection method with respect to user input, the impact of occlusions, performance for different color varieties and lighting conditions. We observed that except for extreme situations (Fig.~\ref{fig:Conditions} rightmost image) most of the difficulties caused by occlusions and lighting conditions can be eliminated with the help of user supervision. Our algorithm achieves $F_1$-score of $.95 - .97$ for different color varieties and lighting conditions outperforming all existing methods.

\begin{figure}[!hbpt]
\includegraphics[width=.58\columnwidth]{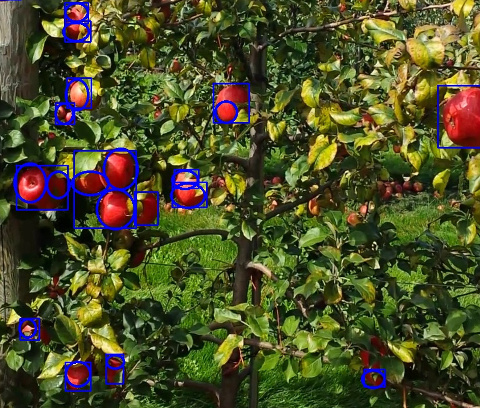}
\includegraphics[width=.32\columnwidth]{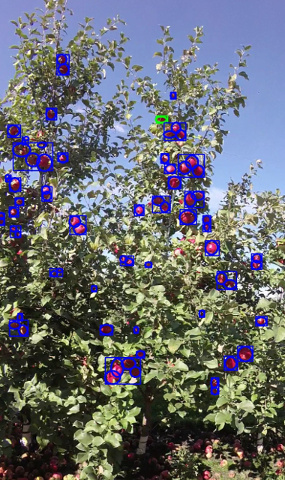}
\caption{Sample output of our algorithm after the counting step. These two images show the difference in the number of pixels occupied by individual apples. Our algorithm works for both cases regardless of any tuning, whereas techniques such as Circular Hough Transform (CHT) need to be tuned for maximum/minimum fruit size for each individual case.}
\label{fig:full_count}
\end{figure}

\begin{figure}[!hbpt]
{\centering
\includegraphics[width=0.3\columnwidth]{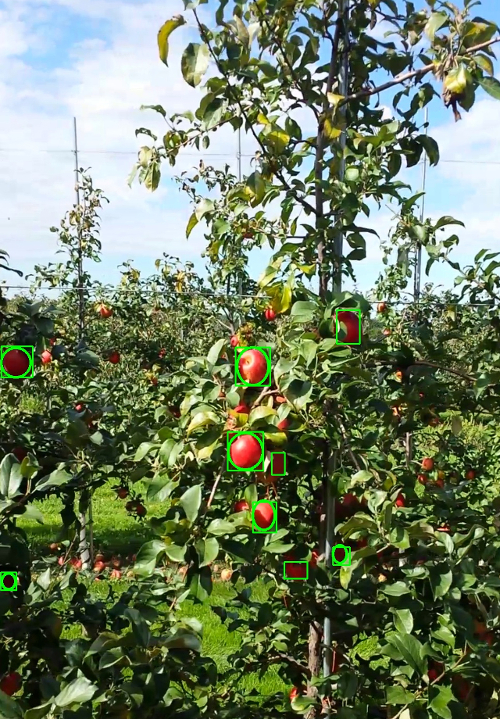}
\includegraphics[width=0.3\columnwidth]{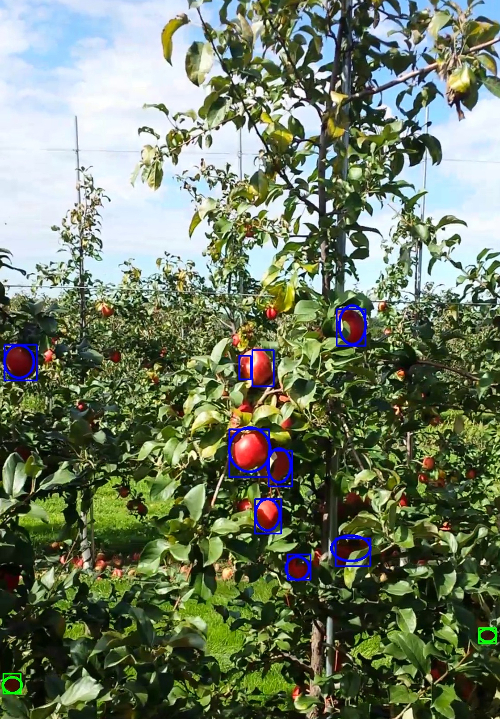}
\includegraphics[width=0.3\columnwidth]{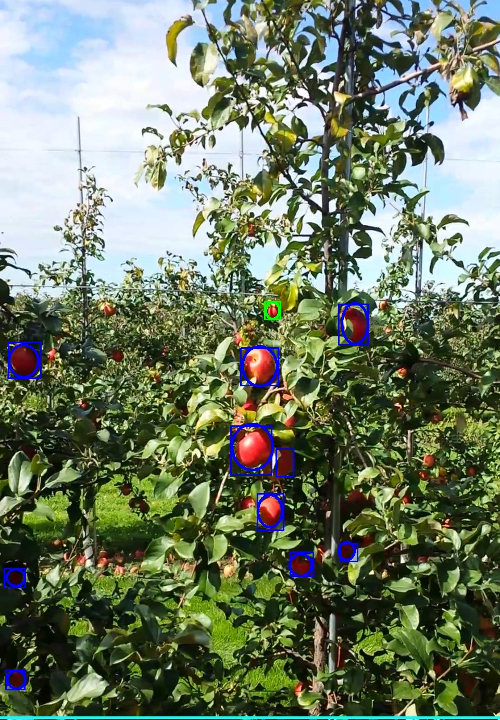}
\caption{
Tracking apples across three consecutive images. Leftmost image is the first frame. In consecutive frames, apples detected earlier are shown in blue. New apples are circled green. \label{fig:tracking}}
}
\end{figure}

 2) We present a novel method for counting the apples from clusters with arbitrarily complex geometry and merge the apple counts across multiple frames utilizing camera motion. Counting apples from a running sequence of images is difficult as apples can be found in arbitrarily shaped clusters in which almost all apples overlap with each other (Fig.~\ref{fig:counting_ex}). Furthermore, because of specularities as well as occlusions due to leaves and branches,  some apples are not detectable at all. We present a method for counting apples based on a classic clustering technique: Gaussian Mixture Models (GMM) (\cite{em}). Our method provides both the counts and location of individual apples in an input image. We model each apple with a Gaussian probability distribution function (pdf) and each apple cluster with a mixture of Gaussians. We present a novel heuristic to find the correct number of components (i.e the number of apples) in the mixture model. Additionally, apples detected in individual frames are tracked across frames to obtain accurate counts (Fig.~\ref{fig:tracking}). We merge the information obtained from the per frame operations across multiple frames utilizing camera motion (approximated by pairwise homography). We validate our counting performance both individually and coupled with tracking. We observe that the accuracy of the per-frame counting method is $94.4\%$. When coupled with tracking, we achieve a varying accuracy of $89\% - 98\%$ for seven different videos. 
 
One of the main takeaways from this paper is that the number of visible apples from a single side of a row varies a lot from datasets to datasets ($40.85\% - 79.83\%$) (See section~\ref{subsec:count_res}). To obtain a consistent estimate of fruit counts, it is essential to have a coherent geometric representation of the entire tree row. With the help of our recent work (\cite{dong2018treeSBA,techreportroy}) we report merged apple counts from both sides of fruit tree rows with a varying accuracy of $91.98\% - 94.81\%$ across three different datasets.

The rest of the paper is organized as follows: in Section~\ref{sec:relwork} we discuss relevant works in literature. In Section~\ref{sec:system} we formalize our problem definition and provide a brief overview of our entire computer vision system. Section~\ref{sec:segmentation}, \ref{sec:counting} present our segmentation and counting methods. Section~\ref{sec:expresult} presents the experimental results. Finally, Section~\ref{sec:conc} presents the conclusion and future work. We start with the related works in the next section.

\section{Related Work} \label{sec:relwork}

There has been a significant recent activity for automating yield estimation(\cite{wang, das2015devices, hung2015feature,gongal2016apple}) in apple orchards. While some of these existing works focus on the entire yield estimation systems, others focus on specific components only. In this section, we first discuss complete yield estimation systems and then focus on individual components. Besides apples, our discussion includes yield estimation systems and components for fruits similar to apples such as citrus, pomegranates, tomatoes etc.

\textbf{Complete Yield Estimation Systems:} \cite{wang} presented a complete system for apple yield estimation. Their system used a side facing, wide-baseline vertical stereo rig mounted on an autonomous ground vehicle. The system operated in controlled artificial illumination settings at night. It uses flashlights to illuminate the apples and classified them using HSV color thresholds. In addition to count information, the stereo system extracted fruit sizes. \cite{hung2015feature} presented a feature based learning approach for identification of red and green apples and extraction of fruit count. They used a conditional random field classifier to segment fruits using color and depth data. \cite{das2015devices} developed a sensor suite for extracting plant morphology, canopy volume, leaf area index and fruit counts that consists of a laser range scanner, multi-spectral cameras, a thermal imaging camera and navigational sensors. They used a Support Vector Machine (SVM) trained on pixel color to locate apples in images. \cite{gongal2016apple}  developed a new sensor system with an over-the-row platform integrated with a tunnel structure which acquired images from opposite sides of apple trees. The tunnel structure is used to minimize illumination of apples with direct sunlight and to reduce the variability in lighting. In contrast to these systems, our goal is to develop a general computer vision system to extract the count of apples in an orchard row from a monocular camera. The system is independent of any particular hardware platform and can be used with a various type of robotic and non-robotic systems. In the rest of this section, we will discuss systems developed for tackling specific portions of the yield estimation system.

\textbf{Apple Detection:} The problem of locating fruits from captured photographs have been studied extensively in the literature. Early systems relied on hard color thresholds. The early methods relying on machine learning were targeted to learn these thresholds. \cite{jimenez2000survey} presented a survey of these early computer vision methods for locating fruits on trees. More recent approaches include \cite{zhouGala}, who presented a method that uses RGB and HSV color thresholds to detect Gala apples. \cite{Linker}, used K-nearest neighbor (KNN) classifier along with blob and arc fitting to find red and green apples. The authors avoid specular lighting conditions by capturing images close to sunset. \cite{changyi2015apple} developed an apple detection method that uses backpropagation neural network (\cite{hecht1989theory}) trained on color features. Similar to them, \cite{liu2016method} presented a method for detecting apples at night using pixel color and position as features in neural networks. The earliest work close to our method is (\cite{tabb2006segmentation}) who developed a method for detecting apples from videos using background modeling by global Gaussian mixture models (GMM). The method relied on images collected using an over-the-row harvester platform that provides a consistent background and illumination setting. In contrast, we operate in natural settings, use GMM for unsupervised clustering in every image and classify them using pre-trained models.

\textbf{Deep Learning for Detecting Apples:} Recent advancements in deep learning inspired researchers to apply these techniques for identifying apples. \cite{bargoti2017image} used Multi-Scale Multi-Layered Perceptron network (MLP) for apple and almond detection. \cite{stein2016image} used a similar deep learning technique to identify mangoes. \cite{chen2017counting} used a Fully Convolutional Neural Network (\cite{long2015fully}) for apple segmentation. Though deep learning methods are accurate, they require a large amount of training data [e.g we trained a Fully Convolutional Network (FCN) using the data from (\cite{bargoti2017image}) and the network did not perform well on our data. When we use some of our data for training though, performance improved drastically]. Generating such training data for different varieties of apples, in different lighting conditions are tedious and cumbersome. In contrast, our method generalizes to any variety of apples and different lighting conditions (as long as apples can be distinguished from other objects by color) using a modest amount of user assistance.

\textbf{Registering Fruits Across Images:} Compared to locating fruits in images, the studies on registering fruits across multiple images has been limited. Only the systems with full yield estimation pipelines studied this problem. \cite{wang} used stereo cameras and point cloud alignment (using odometry and GPS) to avoid double counting. They aligned the apples in 3D space and removed the ones which are within five centimeters of a previously registered apple. \cite{hung2015feature} used sampling at certain intervals to remove overlap between images. \cite{das2015devices} used optical flow and navigational sensors to avoid duplicate apples. In our previous work (\cite{roy2016surveying}), we presented a novel method for registering apples based on affine tracking (\cite{Kannade}) and incremental structure from motion (SfM) (\cite{sinha2014multi}). In this work, we are using homography between frames to track fruits and to avoid double counting. 

To obtain an accurate yield estimate, in addition to registering fruits from a single side, we need to register the fruits from both sides of the row. Only \cite{wang} handles this problem and used point cloud alignment using odometry and GPS. In this paper, we utilize a novel technique that utilizes global and local semantic information (ground plane, tree trunks, the silhouette of foliage etc.) for merging the apple counts from both sides (\cite{dong2018treeSBA,techreportroy}). 

\textbf{Counting Fruits:} Most of the yield estimation work described before report counting results. The methods for counting fruits from segmented images are dominated by circular Hough transforms (CHT) (\cite{silwal2014apple,changyi2015apple,liu2016method}). The main bottleneck of using CHT is that the parameters need to be tuned across different datasets. Another issue with this method is occlusion. Due to occlusion many apples are not fully visible and cannot be approximated by a circle. Different from these methods, \cite{senthilnath2016detection} used  K-means, a mixture of Gaussians and Self-Organizing Maps (SOM) to detect individual tomatoes within a cluster. They used the Bayesian Information Criterion (BIC) (\cite{bic}) to select the optimal number of components in these methods. \cite{chen2017counting} used a  neural network for determining the count of the apples. In contrast, we use a Gaussian mixture model to count the apples from registered and segmented apple clusters. Unlike CHT, our method does not require any parameter tuning and can handle arbitrarily complex clusters. Unlike the neural network based methods, it does not require any training. The method yields competitive results compared to most of the state of the art methods.
In this paper, we present a complete computer vision system for yield estimation in apple orchards. We present a novel segmentation method for detecting apples, a tracking method based on homography and a non-parametric counting algorithm. We start with an overview of the entire system in the next section.
\section{Problem Formulation and Overview of Our Computer Vision System}\label{sec:system}
In this section, we formalize our problem definition and present a brief overview of our system. We start with the problem definition.\\
\textbf{Problem Formulation:} Given a set of images from a calibrated monocular camera facing one side of a row in an apple orchard (where it is assumed that the variance in viewing direction is very small), we want to compute the total apple counts from the entire image sequence.
To solve this problem, we proceed in a per frame manner - separate the apple pixels from others in every frame and count the number of apples in each of them. Afterward, we merge the information across multiple frames by utilizing the approximate camera motion from pairwise homographies. We discuss the components of our system briefly in the rest of this section.
\subsection{Segmentation}\label{subsec:syssegmentation}

The segmentation component takes as input a color image for each frame, and outputs a binary mask which marks whether each pixel in the image belongs to the class \texttt{apple} (Fig.~\ref{fig:seg_pipeline}). This component is presented in detail at Section~\ref{sec:segmentation}.

First, the image is over-segmented into SLIC superpixels (\cite{achanta2012slic}), using the LAB colorspace. A single representative color (mean LAB color of the pixels within the superpixel) is assigned to each superpixel. Then superpixels are clustered by color into approximately $25$ color classes. Finally, it is determined for each class whether it describes apples, based on KL divergence (\cite{goldbergerKLdivergence}) from hand-labeled classes. These hand-labeled classes can be obtained from the unsupervised clusters of the first few frames of a particular video, to easily account for current lighting conditions and the color of the particular apple variety at its particular ripeness.

\subsection{Per Frame Counting}\label{subsec:sysperframecounting}

The per frame counting component takes as input the binary segmented mask for each frame, and outputs a set of bounding boxes and associated integers for each frame where each bounding box represents a connected cluster of apples, and the integer is the estimated number of apples in that cluster (Fig.~\ref{fig:counting_ex}, \ref{fig:full_count}). This component is presented in detail at Section~\ref{subsec:perframecount}. 

First, a connected component analysis is performed on the binary apple mask. Each connected component is examined separately, to determine how many apples it contains. We perform a Gaussian Mixture Model (GMM) based clustering to estimate the number of apples contained within the bounding box, as well as their positions.

\subsection{Camera Motion Approximation}\label{subsec:syscammotion}

The camera motion approximation component takes as input the detected SIFT (\cite{sift}) features in the original input images. It computes a pairwise homography using the SIFT feature matches. 
\subsection{Merging the Counts From Multiple Views}\label{subsec:sysmergecount}

The merging component takes as input a sequence of per frame bounding boxes with apple counts, as well as estimated frame-to-frame homographies of the approximately planar scene.  The output is a total count of unique apples seen in the frame sequence (Fig.~\ref{fig:tracking}, \ref{fig:mergecount}). This component is presented in Section~\ref{subsec:mergecount}.

Essentially, it propagates the computed bounding boxes forward and recomputes the counts when a bounding box overlaps with another one.
\section{Apple Segmentation}\label{sec:segmentation}

In the segmentation stage, the goal is to identify pixels which are likely to belong to an apple. Specifically, the segmentation algorithm takes as input an RGB image and produces a binary mask where pixels belonging to apples are marked as ones and all other pixels are marked as zeros.

One approach to segmentation is to simply choose apple pixels based on predetermined thresholds on color values. While this approach has the advantage of simplicity, in field conditions it often fails due to the variability of lighting conditions. In recent years, deep-learning-based approaches such as convolutional neural networks emerged as powerful, general methods for image classification. Recently their extension to Fully Convolutional Networks (FCN) trained for pixel-wise prediction has gained popularity for fruit and crop segmentation (\cite{chen2017counting}). While these models achieve high accuracy, training a network general enough to accommodate variations in light and visibility conditions remains challenging. Further, training FCN’s for each orchard video separately requires a significant amount of human labor involved in generating ground truth label for each apple.

We propose a semi-supervised image segmentation method based on color and shape features of apples. By itself, the segmentation algorithm runs at $5$ to $6$ frames per second on images of size $1920 \times 1080$ and 15 frames per second for images of the size of $640 \times 480$ on a DELL XPS laptop with 16GB RAM and 2GB GPU memory. The training required from the user is minimal. It is assisted by a simple, convenient user interface. As we show in Section~\ref{subsec:seg_res}, the method generalizes well for the cases where the training and testing data are from different portions of an orchard taken on different days. Our method is expected to work for cases where apples are visibly distinguishable from the surrounding vegetation based on the color. Some working and challenging conditions are shown in Fig.~\ref{fig:Conditions}. 

\begin{figure}[!hbpt]
\centering
    \includegraphics[width= \columnwidth]{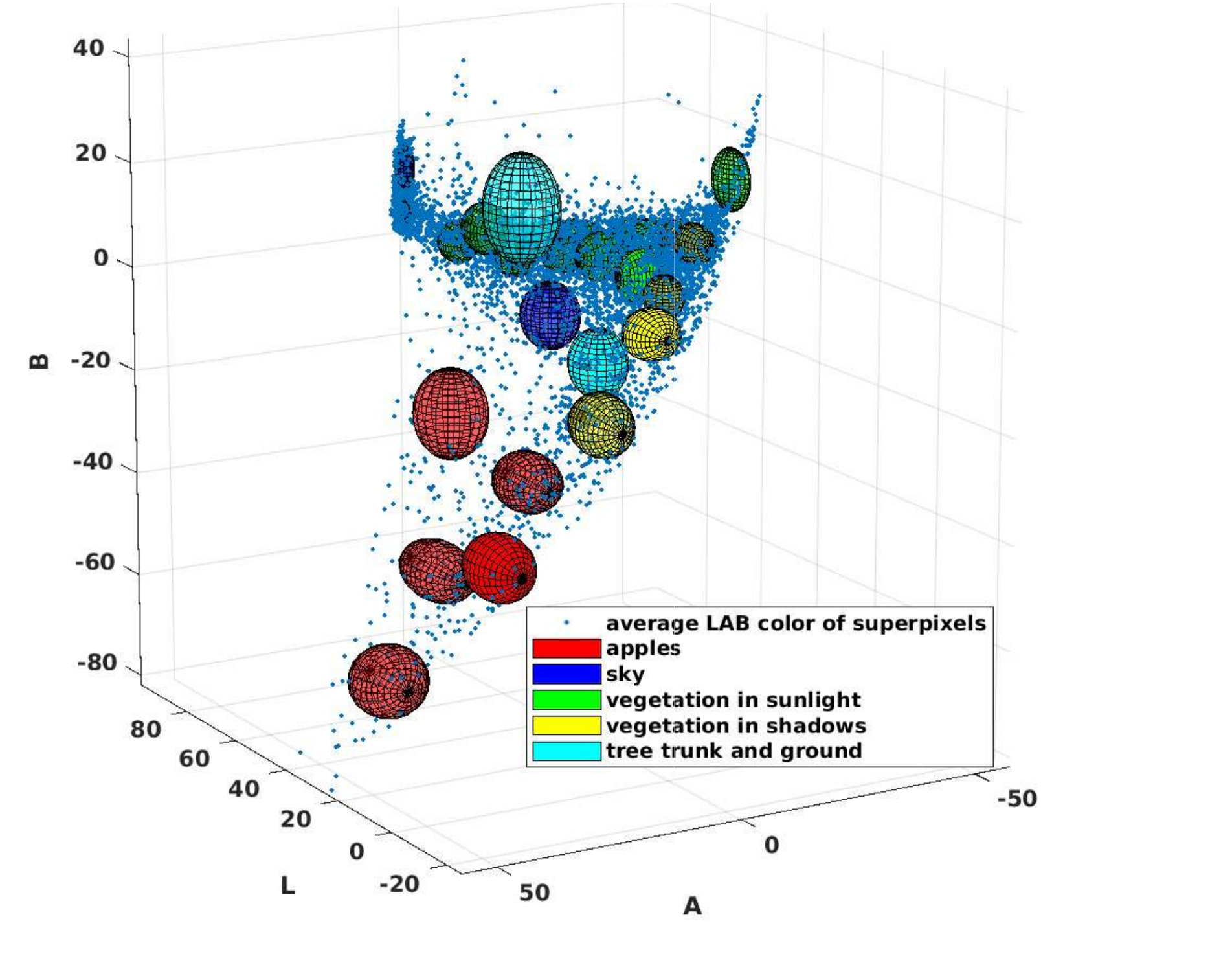}                  
    \caption{Gaussian representation of semantic entities. Each blue data point is the mean LAB color for each superpixel in the image. Each $3D$ ellipsoid represents the Gaussian fit on clusters obtained from unsupervised clustering using EM and GMM. For visualization, showing $19\%$ of the standard deviation for each Gaussian ellipsoid. Red colored ellipsoids are capturing the superpixels belonging to apples.}
    \label{fig:all_semantic_entities}
\end{figure}

\begin{figure}[!hbpt]
\centering
        \begin{subfigure}{0.4\textwidth}
            \includegraphics[scale= 0.4]{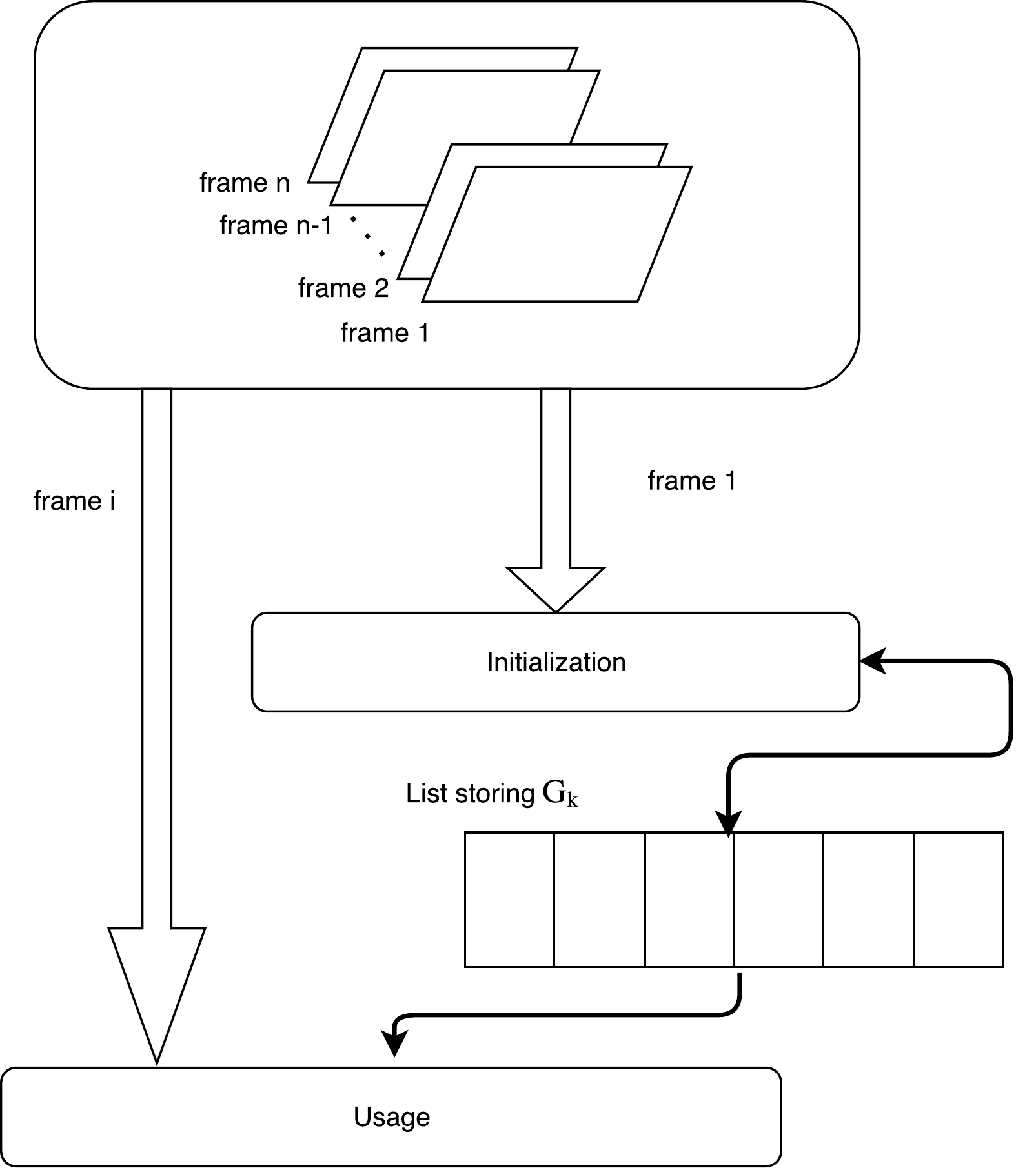}           
            \caption{Input is a video. For each frame in the initialization step, among all Gaussian components $G_i$, $G_k$ are chosen and saved in a list using a UI. These saved $G_k$ are used for finding Gaussian components in the usage step.}
             \label{fig:init_and_usage}
         \end{subfigure}\quad 
         
         \begin{subfigure}{0.35\textwidth}
            \includegraphics[scale= 0.5]{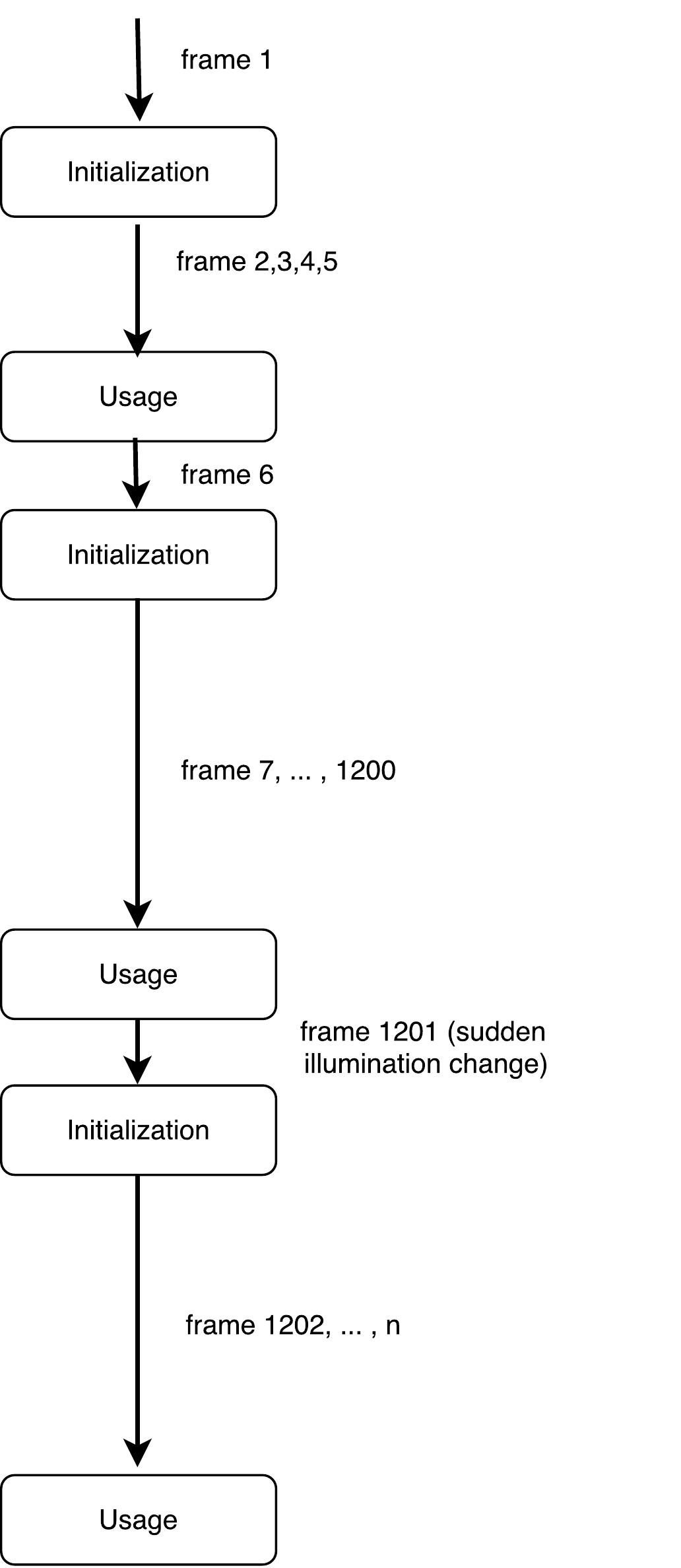}                  
        \hspace*{-1cm}\caption{A general way of using the Usage and Initialization phase for a dataset.}       
        \label{fig:General_paradigm}        
        \end{subfigure}
   \caption{Explanation of the user interface.}
   \label{fig:init_genparadigm}
\end{figure}

\begin{figure*}[!hbpt]
        \centering
        \begin{subfigure}[b]{.23 \textwidth}
            \includegraphics[scale=.15]{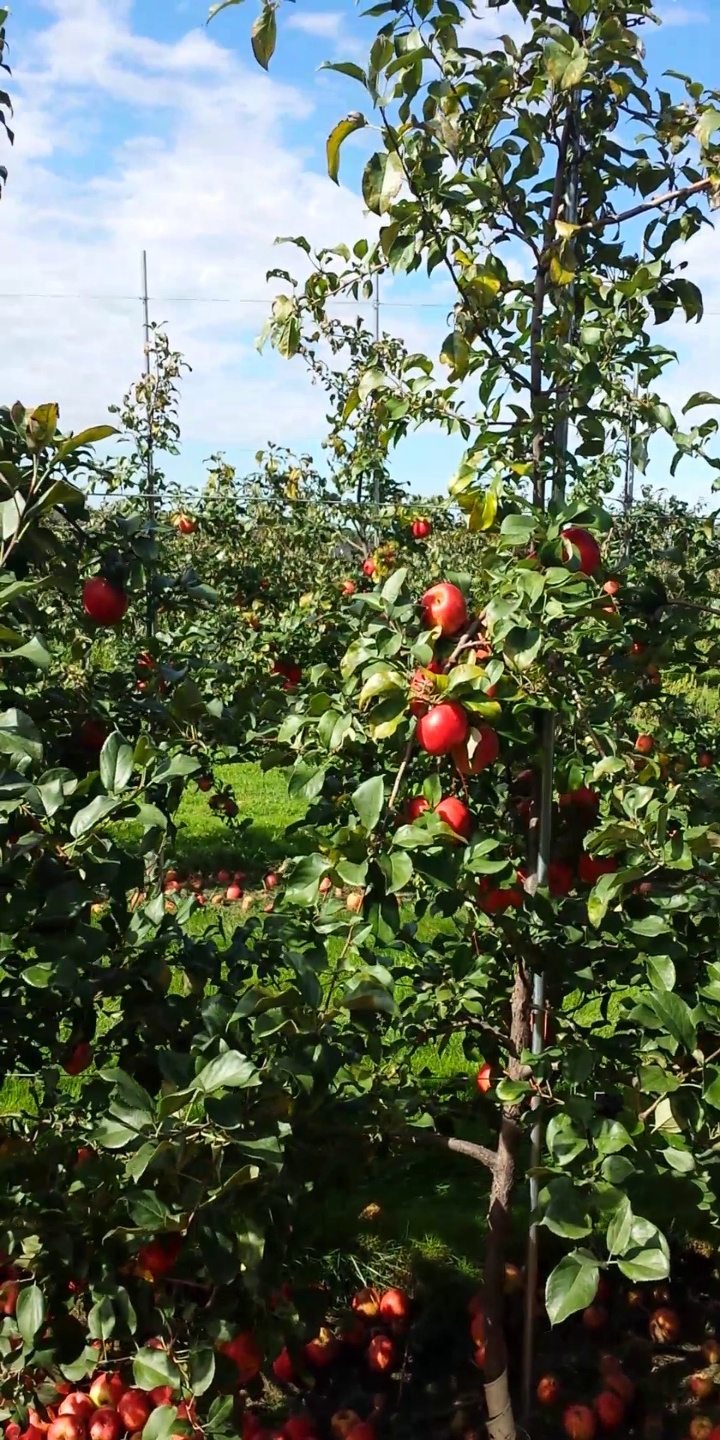}           
            \caption{Input image.}
             \label{fig:pipe1}
         \end{subfigure}\quad 
         \begin{subfigure}[b]{.23 \textwidth}
            \includegraphics[scale=.15]{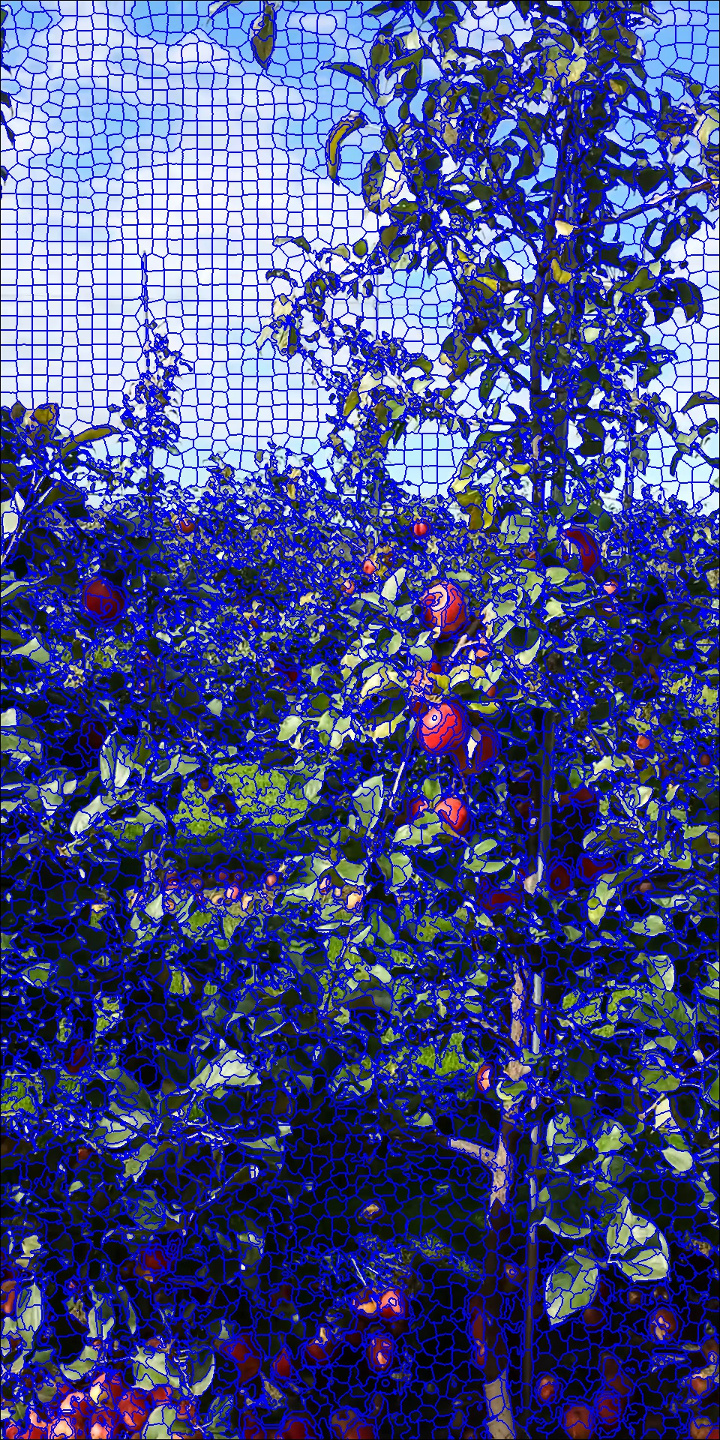}                  
        \caption{Superpixel segmentation.}       
        \label{fig:pipe2}        
        \end{subfigure}
        \begin{subfigure}[b]{.23 \textwidth}      
        
            \includegraphics[scale=.15]{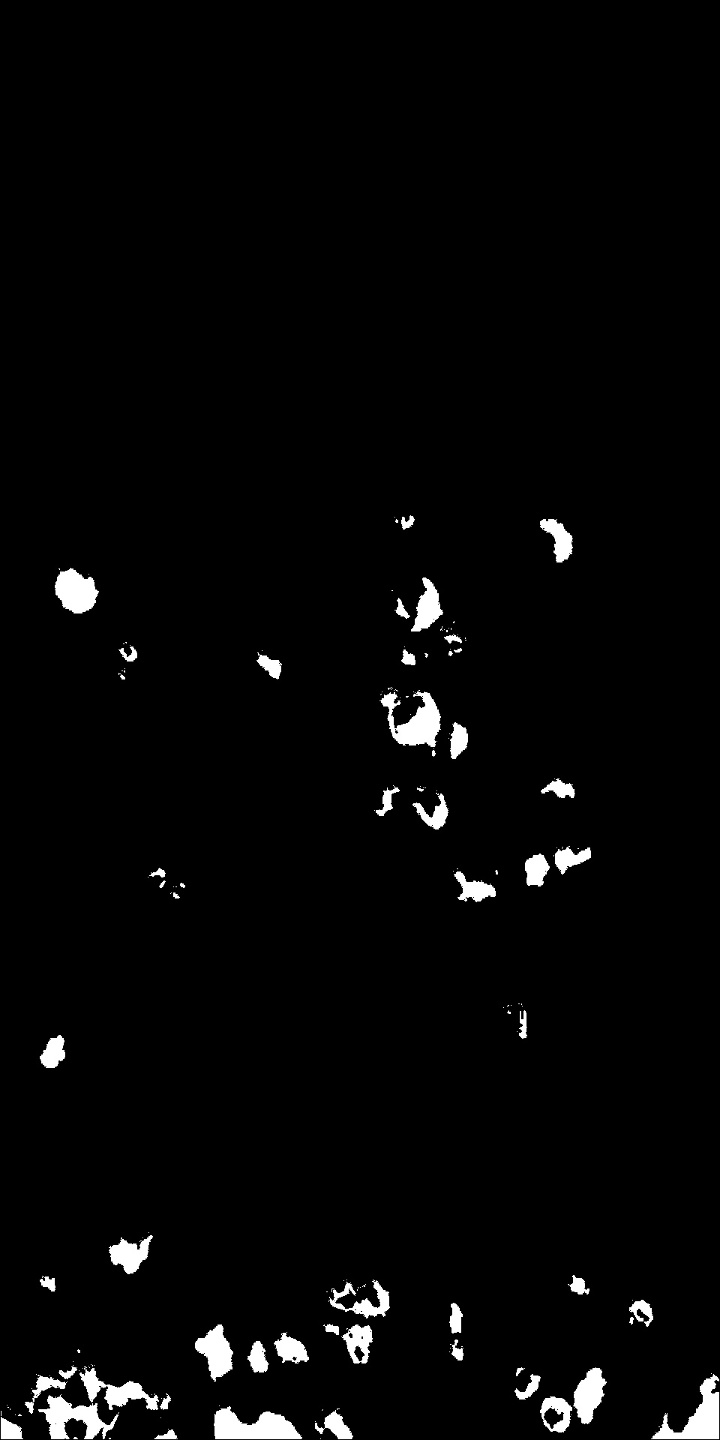}                  
        \caption{Apple segmentation.}       
        \label{fig:pipe4}        
        \end{subfigure}
          \begin{subfigure}[b]{.23 \textwidth}
            \includegraphics[scale=.15]{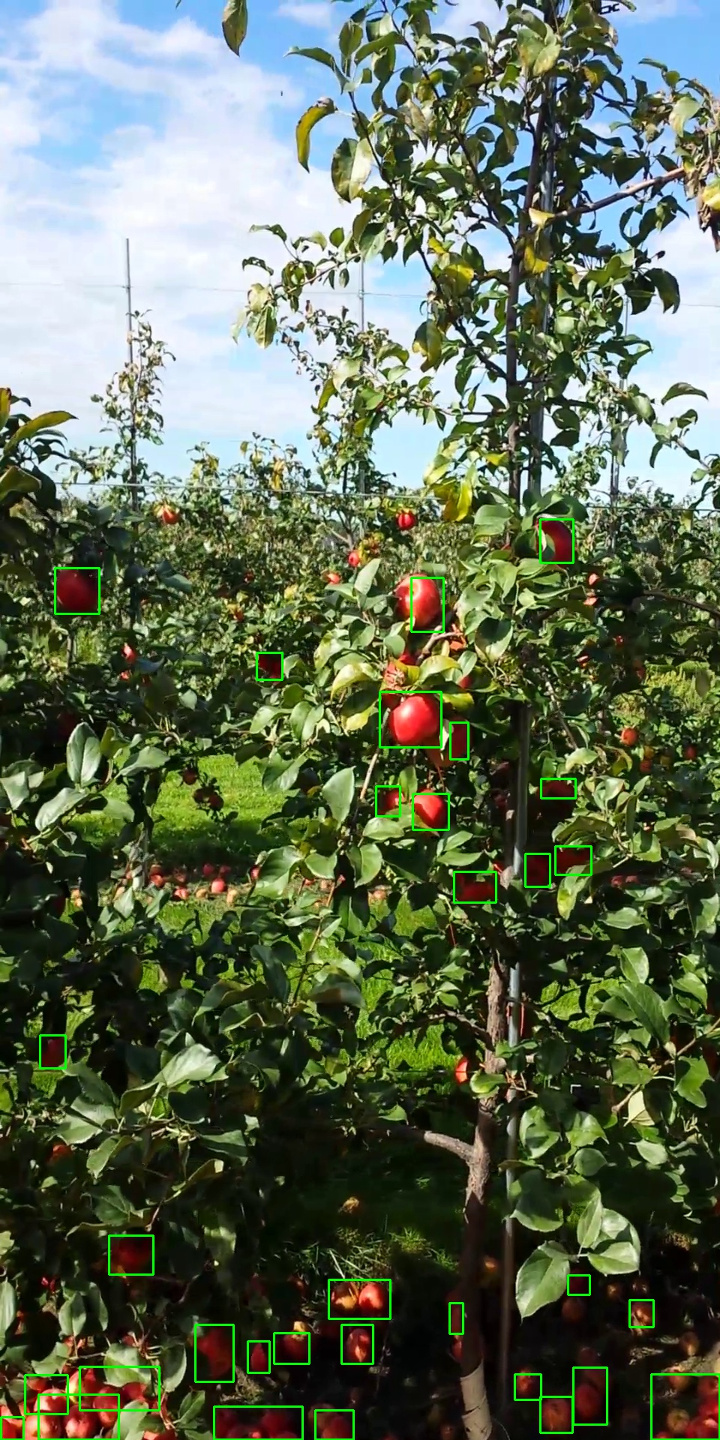}                  
        \caption{Detection.}       
        \label{fig:pipe5}        
        \end{subfigure}
   \caption{Segmentation pipeline.}
   \label{fig:seg_pipeline}
\end{figure*}

\textbf{Details of the segmentation process:} For each frame in the image sequence, we convert it from RGB to LAB (\cite{gauch1992comparison}) color space and perform SLIC superpixel segmentation (Fig.~\ref{fig:seg_pipeline}(\subref{fig:pipe2})) (\cite{achanta2012slic}). This generates a set of superpixels $\mathcal{S}$ for each image.

$$\mathcal{S}=\{\mathbf{s_1},...,\mathbf{s_N}\}$$ 

Here each superpixel $\mathbf{s_i}$ is represented by $\{ \mu^L_i, \mu^A_i, \mu^B_i \}$, the mean L,A,B values for all pixels $\mathbf{s_i}$. We assume that the set $\mathcal{S}$ of superpixels can be modeled as a density function $p(\mathbf{s}|\theta)$ governed by the set of parameters $\theta$. For soft segmentation, we model $\mathcal{S}$ as a Gaussian Mixture Model(GMM)  (\cite{chuang2001bayesian, ruzon2000alpha}) with $M$ components. Hence $p$ is represented by a set of Gaussian components $G_i$ and parameters $\theta$ as the respective mean $\mu_i$ and covariances $\Sigma_i$ for each $G_i$. 

$$ p(\mathbf{s}|\theta)=\sum_{i=1}^M{\alpha_i G_i(\mathbf{s}|\mu_i, \Sigma_i)} $$ such that $\sum_{i=1}^M{\alpha_i}=1$

The likelihood function of the parameters can be written as:

$$ \mathcal{L}(\mu,\Sigma|\mathcal{S}) = \prod_{i=1}^N{p(\mathbf{s_i}|\mu_i,\Sigma_i)} $$

The resultant Gaussian mixture density estimation problem is:

$$(\mu^*, \Sigma^*) = \arg\!\max_{\mu, \Sigma} {\mathcal{L}(\mu,\Sigma|\mathcal{S})} $$

where $\mu^*={\mu^*_1,...,\mu^*_M}$ and $\Sigma^*={\Sigma^*_1,...,\Sigma^*_M}$. Expectation Maximization (EM) is used (\cite{em}) to estimate $\mu^*_i, \Sigma^*_i$ for each Gaussian cluster $G_i$. Each Gaussian cluster $G_i$ thus generated contains similar colored superpixels (Fig.~\ref{fig:all_semantic_entities}) and represents different semantic entities of an orchard frame such as the sky, soil, apples, leaves, branches, etc. 

Next, we identify which among these $G_i$ capture superpixels belonging to apples. We divide this step into two parts: (1)~$\textit{initialization}$ and (2)~$\textit{usage}$ (Fig.~\ref{fig:init_genparadigm}(\subref{fig:init_and_usage})). In the $\textit{initialization}$ phase, we provide an interface for a user to interact with the $G_i$. Here the user is allowed to:
\begin{itemize}
\item select components $G_k$ which completely capture apples. For each selected $G_k$, their respective $\mu_k^*$ and $\Sigma_k^*$ are pushed in a list in memory.
\item delete components from the list stored in memory. This step is generally needed where there is a sudden illumination change and the previous stored components $G_k$ become bad (Fig.~\ref{fig:init_genparadigm}(\subref{fig:General_paradigm})). We can update the old list, and continue the segmentation process from the current frame.
\end{itemize}

In the $\textit{usage}$ phase, the user interface is not invoked. To find components belonging to apples, we perform a simple matching between the current frame's $G_i$ and the saved $G_k$. We use KL Divergence as the distance measure for comparing Gaussians from $G_i$ and $G_k$. For a matched Gaussian $G_i$, all the superpixels within the $90\%$ confidence bounds are identified as superpixels belonging to apples. A normal usage paradigm is shown in Fig.~\ref{fig:init_genparadigm}(\subref{fig:General_paradigm}).

\section{Counting Apples and Merging the Counts across Multiple Frames}\label{sec:counting}
After the segmentation phase, we count the number of apples from each segmented frame and merge the counts across multiple frames utilizing the estimated camera motion. In this section, we present both of these methods in detail. We start with per frame counting in the following section.

\subsection{Per Frame Counting}\label{subsec:perframecount}
Given a segmented input image, we would like to find the number and location of all the apples in the image. We use a Gaussian Mixture Models (GMM) based clustering method. Instead of color, we now focus on the spatial components of the image. This method holds numerous advantages over the Circular Hough Transform (CHT) based techniques (\cite{cht}) - it does not require manual parameter tuning, can handle a significant level of occlusion and find apples of rapidly varying size. A preliminary version of this algorithm was presented in (\cite{roy2016counting}). In ~\cite{roy2016counting}, we showed the comparison of this method with a baseline method similar to CHT and it outperformed it significantly (the accuracy of baseline method $69\%$, the accuracy of GMM based method $91\%$). In this, paper we further validate the method with ground truth from multiple datasets and implement it in a close to a real-time system performing at 2-3 fps. 

In our method, each apple is modeled by a Gaussian probability distribution function (pdf) and apple clusters are modeled as a mixture of Gaussians. We start by converting the input cluster image $I$ to binary. Let this binary image be denoted by $I_{b}$. The locations of the non-zero pixels in the binary image are used as input to GMM. 

\noindent Let X represent the set of apples we are trying to find. Then, we can convert our problem to a Gaussian mixture model formulation in the following way:
\begin{equation}
P(I_b|X) = G^k(\phi,\mu,\Sigma) = \sum_{i = 1}^k \phi_{i} G_{i}(\mu_{i},\Sigma_{i})
\end{equation}

Here, $G^k(\phi,\mu,\Sigma)$ is a Gaussian mixture model with  $k$ components, and $G_{i}$ is the $i$~th component of the mixture. $\mu_{i}$ and $\Sigma_{i}$ are the mean and covariance of the $i^{th}$ component. The covariance matrix $\Sigma_{i} = \left[\sigma_{x_{i}}^2,\sigma_{y_{i}}^2\right]$ is diagonal. $\phi_{i}$ is the weight of the $i^{th}$ component where $\sum_{i= 1}^k \phi_{i} = 1$ and $0\leq \phi_{i}\leq 1$. 

Given model parameters $\theta = \{ \phi, \mu, \Sigma \}$, 
the problem of finding the location of the center of the apples and their pixel diameters can be formulated as computing the world model which maximizes $P(I_b|X)$.

Each component $G_{i}(\mu_{i},\Sigma_{i})$ of the mixture model represents an apple with center at $\mu_{i}$, equatorial radius $2\sigma_{x_{i}}$ and axial radius $2\sigma_{y_{i}}$.

\noindent A common technique to solve for $\arg \max P(I_b|X)$ is the expectation maximization (EM) algorithm (\cite{em}). As is well-known, EM provides us a local greedy solution to the problem. Since EM is susceptible to local maxima, initialization is very important. We used K-means++ (\cite{kmeans}) (which uses randomly-selected seeds to avoid local maxima) for initialization of EM. 

\begin{figure}[!hbpt]

        \centering        
            \includegraphics[scale = .25]{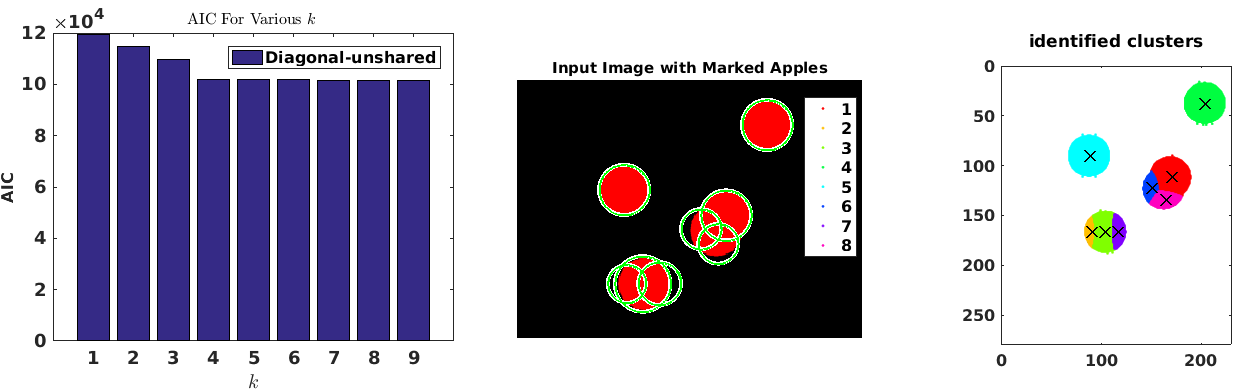}           
        
   \caption{Popular methods like AIC/BIC do not work out of the box for our purposes. These criteria have a tendency of choosing higher values of $k$. In this synthetic image, we have only five circles but the AIC based criterion value (Bar plot in the middle) was lowest for $k = 6$ and consequently, it finds eight apples.}
   \label{fig:aic}   
\end{figure}

\textbf{Selecting the Number of Components:} In our problem formulation, the number of components $k$ is the total number of apples in image $I$. EM enables us to find the optimal location of the apples given the total number of apples $k$. Our main technical contribution is a method to calculate the correct $k$. Let the correct number of apples in the input image be $\kappa$. We tried different state-of-the-art techniques (Akaike Information Criterion (AIC) (\cite{mdt}), Minimum Description Length (MDL) (\cite{mdt}) etc.) for finding $\kappa$. None of them worked out of the box for our purposes (Fig.~\ref{fig:aic}). Therefore, we propose a new heuristic for evaluating mixture models with a different number of components based on MDL. 

Let  $\sigma_{min} = min(\sigma_{x_{i}},\sigma_{y_{i}})$ and $\sigma_{max} = max(\sigma_{x_{i}},\sigma_{y_{i}})$. Using the mean and covariances of the $i$th component we define a $2D$ Gaussian kernel $\mathcal{G}(\mu_{i},\sigma_{max})$ where  $\sigma_{max}$ is the variance. Let $P(\mu_{i})$ denote the response of the kernel when placed at the center $\mu_{i}$ in the original input image $I$ and $C_{i}$ denote the total number of pixels clustered by $G_{i}(\mu_i,\Sigma_i)$. For each component $G_{i}(\mu_i,\Sigma_i)$, of the mixture model $G^k(\phi,\mu,\Sigma)$ we define the reward $R_{i}$ in the following way,

\begin{equation}
\begin{split}
R_i(G_{i}) =  \phi_{i}\left[ P(\mu_{i})+  P(\mu_{i})\left(\frac{\sigma_{min}}{\sigma_{max}}\right)^2 +\right. \\ \left.  P(\mu_{i})\frac{C_{i}}{\pi \sigma_{max}\sigma_{min}} - \frac{1}{3}\left( \pi\sigma_{x_{i}}\sigma_{y_{i}} -C_{i} \right) \right]
\end{split}
\label{eq:reward}
\end{equation}

For most of the images, we only capture the frontal views of the apples, which can be easily approximated by circles lying on a plane. All four terms in equation~\eqref{eq:reward} reward specific spatial characteristics of the Gaussian pdf related to this fact. $P(\mu_{i})$ represents the strength of the distribution in terms of pixel values and is present in the first three terms. The second term rewards circular shaped distributions using the eccentricity of the pdf. As the eccentricity $\epsilon = \sqrt{1- \frac{\sigma_{min}^2}{\sigma_{max}^2}}$ for circles is zero, we use $1-\epsilon^2 = \left(\frac{\sigma_{min}}{\sigma_{max}}\right)^2$ as the rewarding factor. The third term rewards coverage. The fourth term penalizes Gaussian pdfs covering large area and clustering very few points.

Now if we find out the reward $R_i(G_{i}(\mu_i,\Sigma_i))$ for all the components $k$, the total reward for the mixture model $G^k(\phi,\mu,\Sigma)$ can be computed by summing them together.

Next, we define the penalty term. The traditional MDL penalty term is $U = c p \log(|Y|)$ where $p$ is the number of parameters in the model, $|Y|$ is the total size of the input data, and $c = \frac{1}{2}$ is a constant. Based on this principle, our penalty term is  $V(G^k(\phi,\mu,\Sigma))$ is defined as the following
\begin{equation}
V(G^k(\phi,\mu,\Sigma)) = c'(3k) \log(\sum_{\vec{x}}(I_b(\vec{x}) \neq 0)))
\label{eq:penalty}
\end{equation}
where $x$ represents the pixel index across the image $I_b$. Compared to the traditional MDL based penalty we have the constant $c' =\frac{3}{2}$ instead of $c =\frac{1}{2}$. This is attributed to the fact that the reward expression~\eqref{eq:reward} has three terms compared to one. The number of components $k$ is multiplied by three as each Gaussian has three parameters $\left[\mu_i,\sigma_{x_i}, \sigma_{y_i}\right]$. With these terms defined, we choose the correct number of components $\kappa$ in the following way:

\begin{equation}
\kappa = \argmax_k R(G^k(\phi,\mu,\Sigma))- V(G^k(\phi,\mu,\Sigma))
\label{eq:numcomp}
\end{equation}

\begin{figure}[!hbpt]
\begin{subfigure}{\columnwidth}
        \centering        
            \includegraphics[width = \columnwidth]{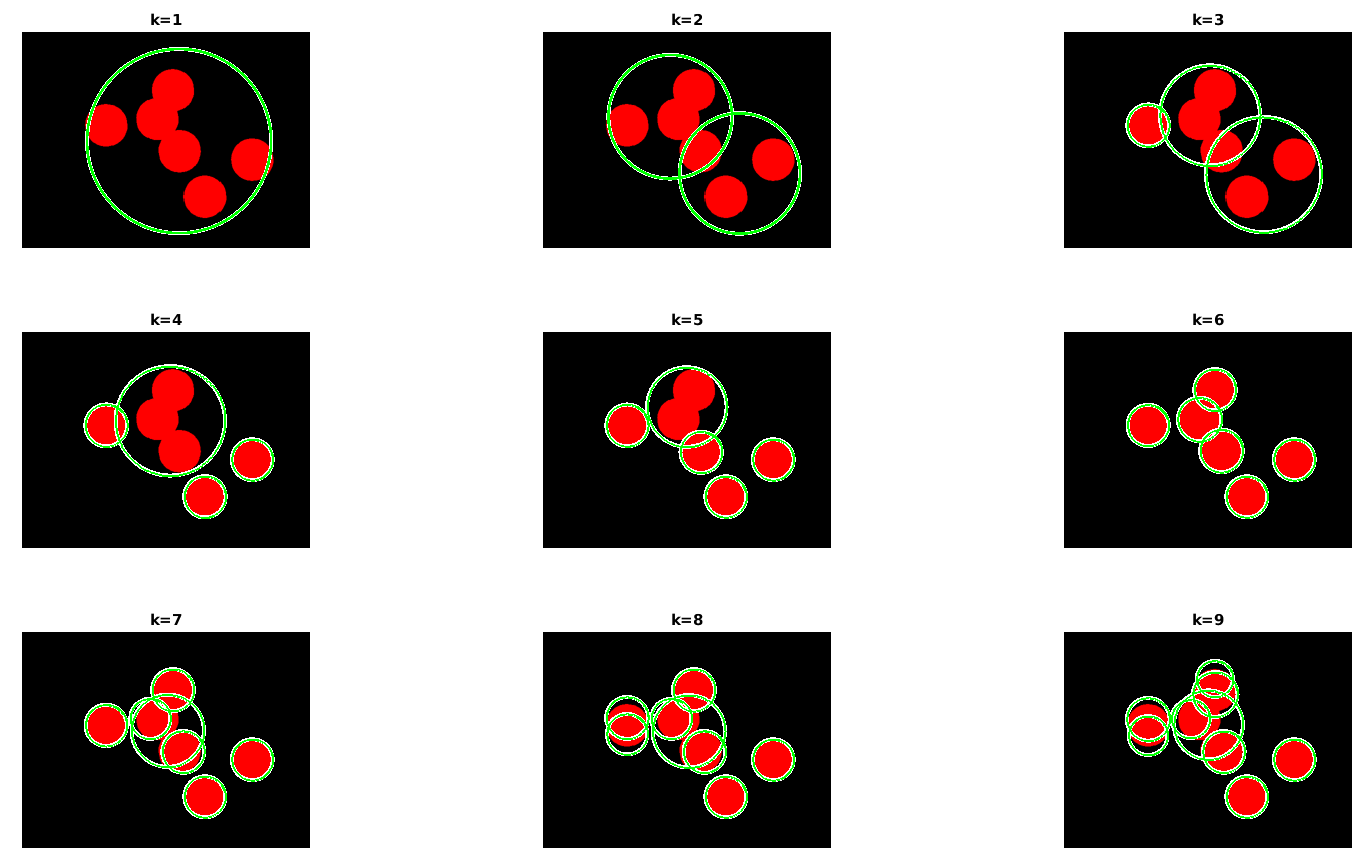}           
        
   \caption{Predicted circles for different number of components in GMM.}
   \label{fig:gmmsyn}
   \end{subfigure}\\ \begin{subfigure}{\columnwidth}
   \centering
   \begin{subfigure}{.32\columnwidth}
   \centering
            \includegraphics[width = \textwidth]{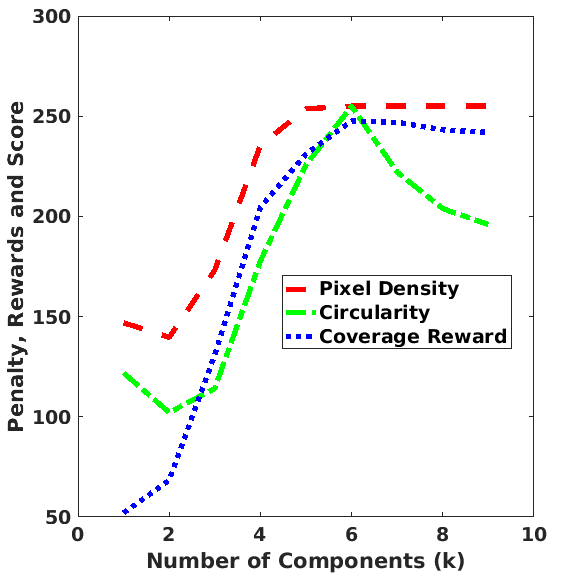} 
            \caption{Rewards.}
        \label{fig:gmmkplotrew}               
       \end{subfigure}\begin{subfigure}{.32\columnwidth}  
       \centering
       \includegraphics[width = \textwidth]{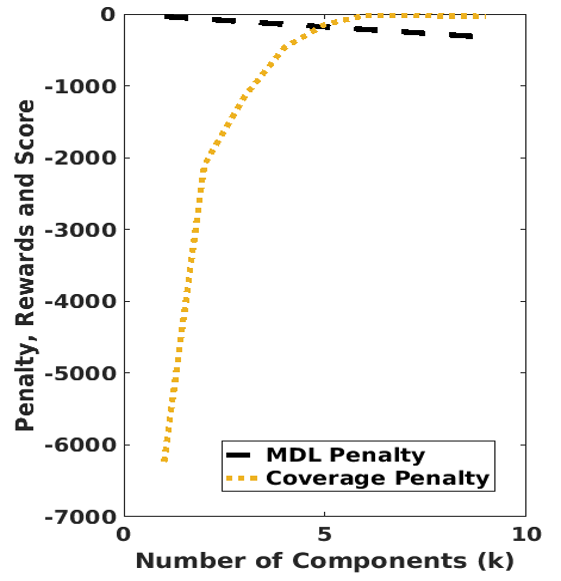} 
            \caption{Penalties.}
        \label{fig:gmmkplotpen}    
       \end{subfigure} \begin{subfigure}{.32\columnwidth}   
       \centering
       \includegraphics[width = \textwidth]{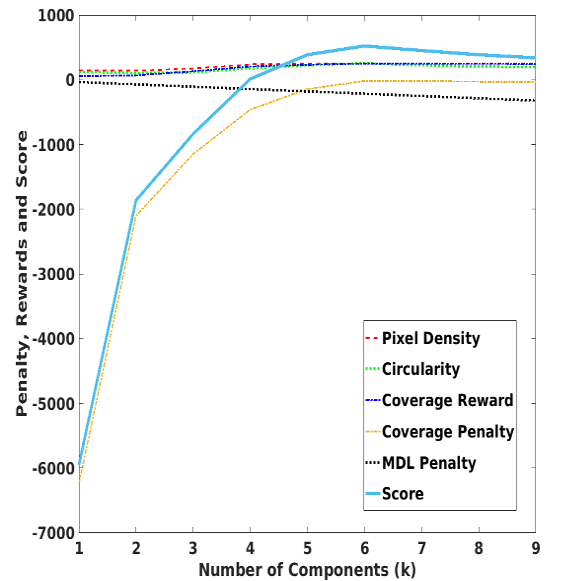} 
            \caption{Final score.}
        \label{fig:gmmkplot}   
       \end{subfigure}
       \end{subfigure}
  \caption{A synthetic example consisting of six random circles and plots illustrating how the number of components are selected. Figure~(\subref{fig:gmmsyn}) shows how the pdfs cover the circles for different values of $k$. Figure~(\subref{fig:gmmkplotrew})-(\subref{fig:gmmkplot})shows the score calculated from the rewards and penalties following the right hand side of equation~\eqref{eq:numcomp}. The plot shows that the score is maximum for $k = 6$ which is indeed the correct number of components.}
   \label{fig:grpsyn}
\end{figure}    

\begin{figure}[!hbpt]

        \centering        
            \includegraphics[scale = .2]{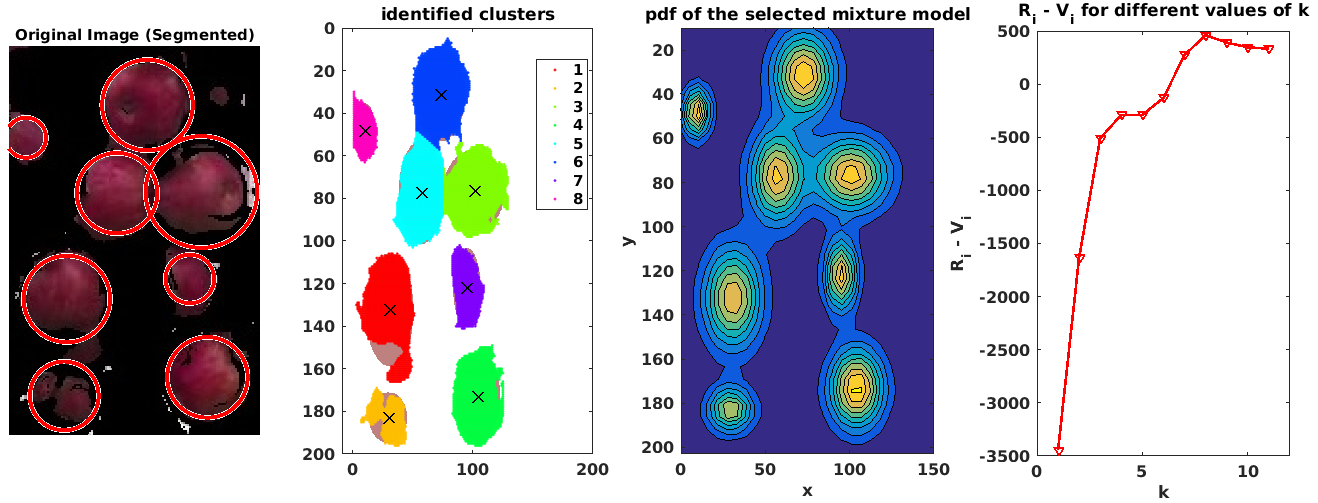}           
        
   \label{fig:gmmex}  
  \caption{A sample output from GMM on real data.}
   \label{fig:grpcongmm}
\end{figure}

To have a better understanding of the selection procedure, we demonstrate a synthetic example at Fig.~\ref{fig:grpsyn}(\subref{fig:gmmsyn}). From Fig.~\ref{fig:grpsyn}(\subref{fig:gmmkplotrew}), it is evident that except for $k =6$, other mixtures have low circularity. The coverage rewards and pixel density components increase with $k$ and converge to a steady state. While the penalty for minimum description length principle increases with $k$, generally the penalty for coverage decreases with $k$. In this example, the crucial factors in determining the score are circularity and the coverage penalty. For $k = 6$ circularity is at the peak and coverage penalty is lowest and consequently, the score was maximum for $k = 6$. The plot of the corresponding rewards, penalties and final scores are shown in Fig.~\ref{fig:grpsyn}(\subref{fig:gmmkplotrew}) - (\subref{fig:gmmkplot}). We show sample results from our datasets in Fig.~\ref{fig:grpcongmm}. 

After executing the counting algorithm on each of the bounding boxes we have the location of the apples within each box and the number of apples per box. This is the input to our merging method that merges the apple counts across multiple frames. We describe this procedure in details in the next section.

\subsection{Merging Count Across Multiple Frames}\label{subsec:mergecount}

The per-frame counting method provides us with the apple counts for a single frame. In natural settings though, the visibility of a particular cluster can change drastically with a change in camera position. A cluster might be completely invisible/partially visible from a particular view and yet clearly visible from other views. For these reasons, our segmentation method may not able to obtain the perfect segmentation in each frame and consequently, the predicted number of apples might be incorrect. Therefore, to obtain the correct apple count for each cluster we merge the counts from different frames. For this operation, we need to establish the correspondence between the clusters across multiple frames. It is executed by utilizing the camera motion. 

\begin{figure}[!hbpt]
        \centering
            \includegraphics[width =0.99\columnwidth]{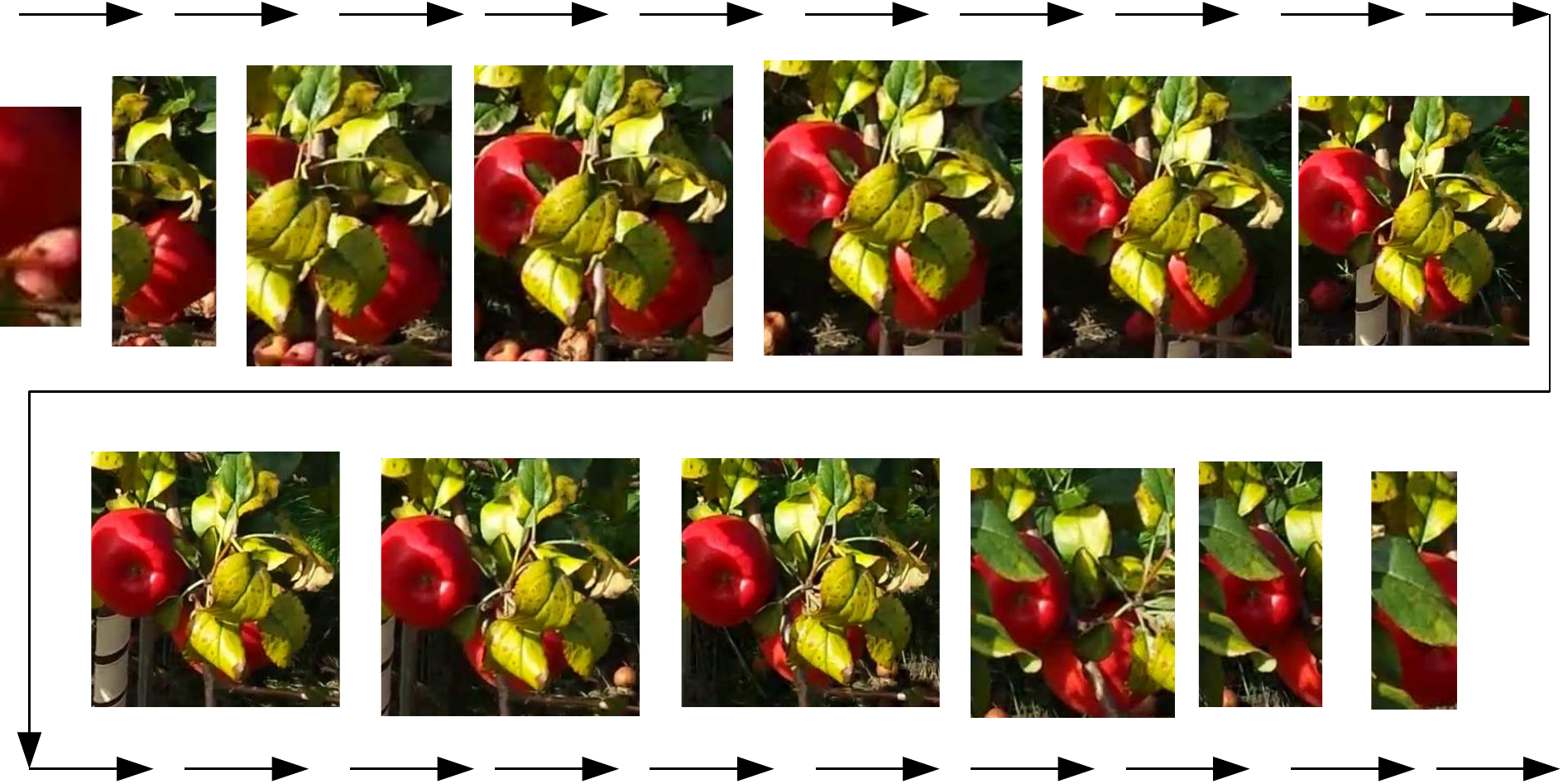}           
   \caption{Tracking an apple cluster over $31$ frames. For clarity only $13$ frames with significant changes are shown. The arrows indicate the direction of the views from start to end.}
   \label{fig:trackcluster}
\end{figure}

As our camera viewing direction does not change much, and the scene is roughly planar, we model the camera motion between consecutive frames by pairwise homography (\cite{eshel2008homography}). The homography between frames is estimated by matching SIFT (\cite{sift}) features above the ground. Using homography we will keep track of the boxes generated by the connected component analysis across the entire image sequence (Fig.~\ref{fig:trackcluster}).

Let $b^1_1, b^1_2, \ldots b^1_m$ be the bounding boxes generated by connected component analysis for frame $1$. Let the apple count for each of this boxes be $c^1_1, c^1_2, \ldots c^1_m$ (computed by the counting method). When we find a bounding box for the first time we initialize a counting list that contains computed counts from the first frame.

Now for frame $2$, let the bounding boxes be $b^{2}_1, b^{2}_2, \ldots b^{2}_n$ and the counts be $c^{2}_1, c^{2}_2, \ldots c^{2}_n$. Let the homography that maps frame $1$ to frame $2$ be $^2 H_{1}$. We propagate all the bounding boxes in frame $1$ to frame $2$ using $^2 H_{1}$. This is executed by multiplying the center of each bounding box with $^2 H_{1}$. Next, we check the overlap between these propagated bounding boxes and the original bounding boxes on frame $2$. If the overlap is more than $10\%$ we assume that these bounding boxes correspond to the same cluster. We will add the apple count to the list initialized previously using the following rules:

\begin{itemize}

\item When a bounding box in the current frame does not overlap with any of the propagated bounding boxes, a new counting list for this box is initialized with the current count.

\item When only one bounding box in the current frame overlaps with a propagated bounding box, the count from the bounding box in the current frame is added to the counting list for the propagated box

\item Otherwise, when two or more bounding boxes overlap with a bounding box from the previous frame, their counts are added to obtain the total count to be recorded in the list. To have a better understanding of this rule, we consider the following scenario: Let $b^2_1, b^2_3$ be overlapping with $b^1_1$. Prior to frame $2$ the counting list of $b^1_1$, $c_{b^1_1}$ had only one entry, $c^1_1$. As there is overlap, a new entry will be inserted. This new entry is $c'^2_1 = c^{2}_1 + c^{2}_3$.

\item The overlapping bounding boxes are unioned to obtain a new bounding box covering all of them. These new bounding boxes will be propagated to the next frame.

\end{itemize}

\begin{figure}[!hbpt]

        \centering        
            \includegraphics[width =0.90\columnwidth]{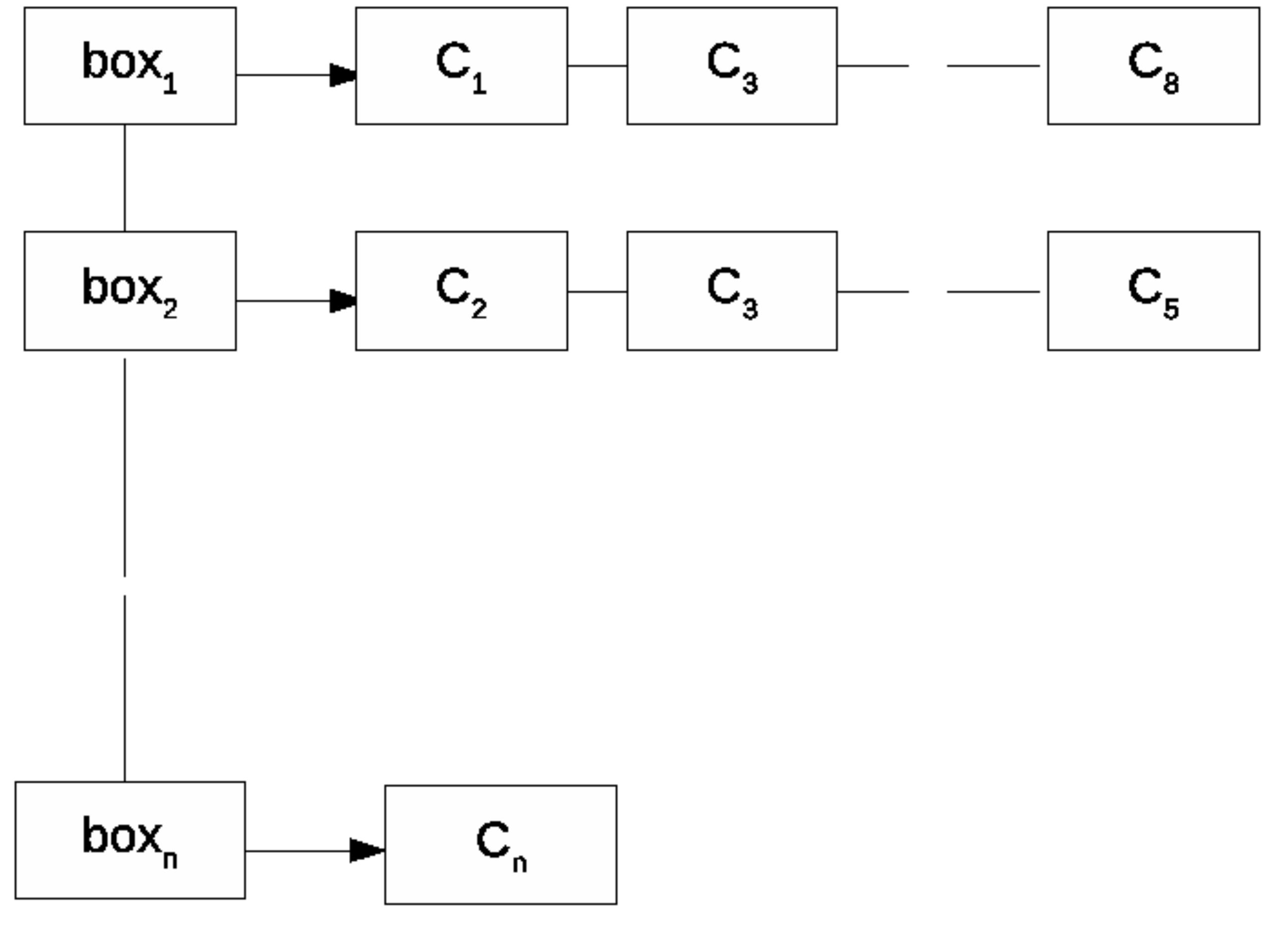}           
        
  
  \caption{Merging Apple Counts Across Frames. We keep track of the corresponding bounding boxes generated connected component analysis across multiple frames using homography. The count for every box from different frames is recorded throughout the entire video. In this example, $box_1$ appeared in frame $1,3, \ldots ,8$ and the counts for each of these frames $C_1,C_3 \ldots C_8$ were recorded. The median counts for each box are reported as the final output of the counting method.}
   \label{fig:mergecount}
\end{figure}    

At the end of the image sequence, we have a set of unique boxes with count lists. We compute the median for each of the boxes and the sum of these are reported as the total count (Fig.~\ref{fig:mergecount}).

\section{Experimental Results}\label{sec:expresult}
In this section, we present experimental results validating our algorithms. We start with the datasets.

\subsection{Datasets}\label{subsec:datasets}

\begin{figure}[!hbpt]
        \centering
        \begin{subfigure}[b]{.45\columnwidth}
            \includegraphics[width=\textwidth]{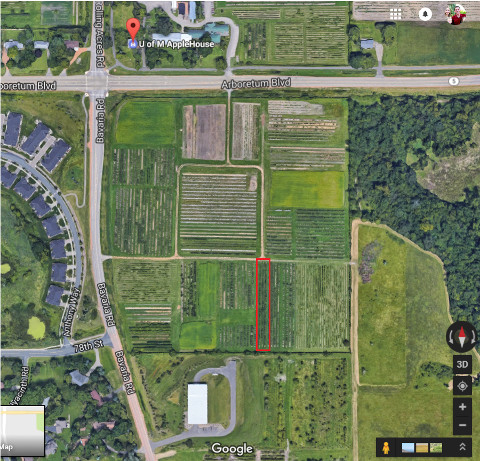}           
       \end{subfigure}\quad
        \begin{subfigure}[b]{.45\columnwidth}
            \includegraphics[width=\textwidth]{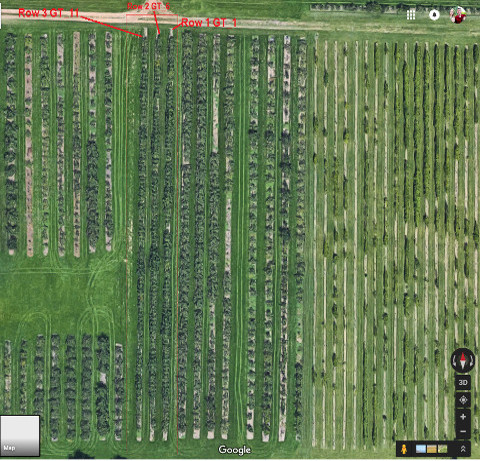}           
        \end{subfigure}
   \caption{Data collection site with annotated rows. The figure on left shows the orchard location and the figure on right shows the individual rows of apple trees.}
   \label{fig:datasite}
\end{figure}

To verify the performance of our algorithms, we collected multiple datasets. We group these datasets into two categories, namely - ``Validation Datasets and  Training Datasets''. As their names suggest, they are used for the purposes of validation, and training. All data were collected at the University of Minnesota Horticulture Research Center at Victoria Minnesota (Fig.~\ref{fig:datasite}), over the course of two years (2015 - 2016). 

Each of the videos collected from these datasets is tagged sunny/shady/cloudy based on the weather condition and whether the particular side of the row (captured in the video) was facing/opposing the sun.

\paragraph{\textbf{Validation Datasets}}
Since this is a research site, it is home to many different kinds of apple trees. We arbitrarily chose four different sections in the orchard. We collected seven videos from these four segments, all of which were annotated manually with fruit locations (annotation procedure will be discussed in the next section). We also collected ground truth for all of them by labeling the apples physically by stickers and measuring their diameter after harvest. The details of these datasets are the following:
\begin{figure*}[!hbpt]
        \centering
        \begin{subfigure}{.18\textwidth}
            \includegraphics[width=\textwidth]{dataset11}           
       \end{subfigure}\quad \begin{subfigure}{.18\textwidth}
            \includegraphics[width=\textwidth]{dataset222}           
        \end{subfigure}\quad \begin{subfigure}{.18\textwidth}
            \includegraphics[width=\textwidth]{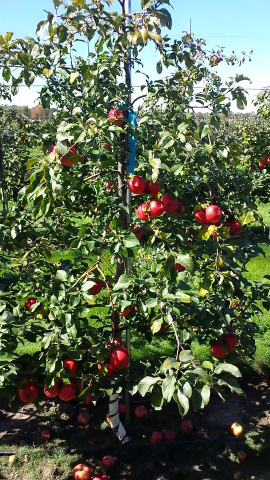}           
       \end{subfigure}\quad \begin{subfigure}{.18\textwidth}
            \includegraphics[width=\textwidth]{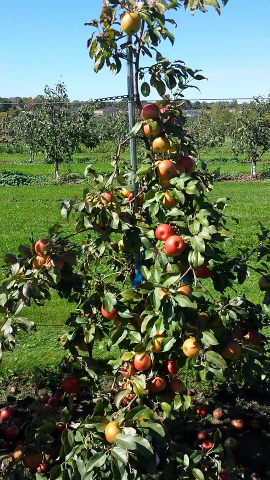}           
       \end{subfigure}\quad \begin{subfigure}{.18\textwidth}
            \includegraphics[width=\textwidth]{dataset20151}           
       \end{subfigure}
   \caption{Sample images from used datasets. First four figures show sample images from the test datasets and the last one shows a sample image from our training dataset.}
   \label{fig:datasets}
\end{figure*}
\begin{description}
\item[Dataset1] This dataset contains six trees. Most of the apples on these trees were fully red and the trees were mostly planar (most of the apples are visible from both sides). We collected videos from both sides (the side facing the sun and opposing the sun) of the row. In total, there were $270$ apples in these six trees. See Fig.~\ref{fig:datasets} (leftmost) for a sample image from this dataset.
\item[Dataset2] This dataset contains four trees. The trees had a mixture of red and green apples and complex (non-planar) geometry. We collected a video from a single side of the row (the side facing the sun). In total, there were $568$ apples in these four trees. See Fig.~\ref{fig:datasets} (second from left) for a sample image from this dataset.

\item[Dataset3] This dataset contains ten trees. Apples in these trees were mostly red and the trees had non-planar geometry. In total, there were $274$ apples in these ten trees. We collected videos from both sides (the side facing the sun and opposing the sun) of the row. See Fig.~\ref{fig:datasets} (third from left) for a sample image from this dataset.

\item[Dataset4] This dataset contains six trees. Fruits in these trees were a mixture of red and green apples and the trees had non-planar geometry. We collected videos from both sides of the row (the side facing the sun and opposing the sun). In total, there were $414$ apples in these six trees. See Fig.~\ref{fig:datasets} (fourth from left) for a sample image from this dataset.
\end{description}

All of the videos collected from these datasets are used to validate our segmentation and counting algorithms. Additionally, Dataset $1,3$ and $4$ are used to verify the total fruit counts from both sides. As we only collected video from a single side of the row from Dataset2, it is not used to verify both side fruit counts.

\paragraph{\textbf{Training Datasets}}
To validate how our algorithms perform without user supervision, we picked a dataset from 2015. This dataset contains $76$ trees. Fruits in these trees were a mixture of red and green apples and the trees had non-planar geometry. We collected a video from a single side of a row (it was a cloudy day, both sides were illuminated similarly). This video is not annotated manually and ground truth for this dataset is unknown. We only used this dataset for developing a model to identify apples without any user intervention. In other words, the user input was provided for the 2015 dataset and used without modification for 2016 dataset.

The four validation datasets were captured using a Samsung Galaxy $S4$ camera in September 2016. The training dataset was collected in 2015 using a Garmin VR camera. It is notable that, our data collection facility is not a commercial orchard. Consequently, there is a great amount of variability in the shape of the trees even within the same crop row, which makes the yield estimation problem harder. In the next section, we look at the process of annotating the datasets.
\subsection{Manual Annotation of Apples for Verifying Detection and Counting}\label{subsec:manuallabel}

In order to validate our segmentation and counting methods, we need image level ground truth (detected apples in each individual input image). This is different from the number of harvested apples from trees. For this purpose, we annotated the boundary of individual apples in the input images. This provides us with the ability to compare the bounding boxes generated by our algorithm with bounding boxes drawn manually.

\begin{figure}[hbt]
        \centering
        \begin{subfigure}{.4\columnwidth}
            \includegraphics[width=\textwidth]{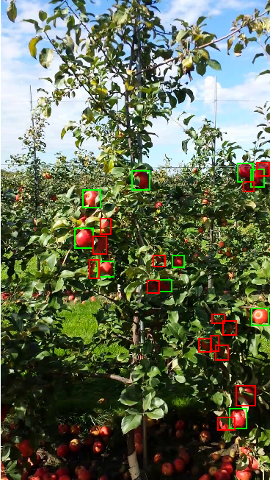}
        \end{subfigure}\quad \begin{subfigure}{.4\columnwidth}
                 \includegraphics[width=\textwidth]{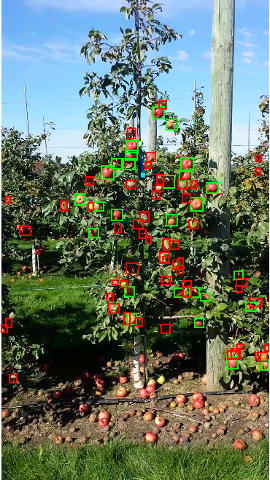}                        
        \end{subfigure}
   \caption{ Manual annotation of clearly visible and marginally visible apples for two sample images. An apple was considered $\textit{clearly visible}$ if more than half of its cross-sectional area and more than half of its perimeter were unoccluded (green boxes). Otherwise, if it was still detected as an apple by the human, it was marked as $\textit{marginally visible}$ (red boxes).}
   \label{fig:goodMarginal}
\end{figure}

For manual annotation, frames were selected arbitrarily every $1-3$ second for the test videos (frame rate $30$ fps), depending on how much the camera moved since the last annotated frame. For each of these frames, apples were tagged as $\textit{clearly visible}$ or $\textit{marginally visible}$ based on visibility guidelines (Fig.~\ref{fig:goodMarginal}). An apple was considered clearly visible if more than half of its cross-sectional area and more than half of its perimeter were unoccluded. Otherwise, if it was still detected as an apple by the human, it was marked as marginally visible. The marginally visible apples have more ambiguous bounding boxes, and might not even have a one-to-one mapping between boxes and apples. In addition to these guidelines, apples that were growing on trees in the rows behind the main row of interest, and apples that had fallen to the ground, were not tagged. 

Seven videos collected from the validation datasets (Dataset1 to Dataset4), described in the previous section, were tagged in this manner. The manually drawn bounding boxes were then propagated to the rest of the frames using camera motion between the frames. In the next sections, we investigate the performance of our segmentation and counting algorithms using these hand labeled datasets. Additionally, we will look at the relationship of computed yield vs actual yield for Dataset1, Dataset2, and Dataset4.

Our segmentation algorithm detects the ground apples as well. The manual annotation procedure ignores them. To remove ground apples, we let users choose a single line at the start of each video. This line is propagated using homography for the entire image sequence and any apple detected below this line are labeled as ground apples, and they are ignored for both segmentation and counting.
\subsection{Performance Evaluation of the Segmentation Method}\label{subsec:seg_res}
\begin{figure*}[!hbpt]
        \centering
        \begin{subfigure}[b]{.31\textwidth}
            \includegraphics[width =\textwidth]{seg_eval}           
            \caption{Evaluation metric for segmentation}
             \label{fig:segmentation_evaluation}
         \end{subfigure}\quad 
         \begin{subfigure}[b]{.30 \textwidth}
         \raggedleft
            \includegraphics[width = \textwidth]{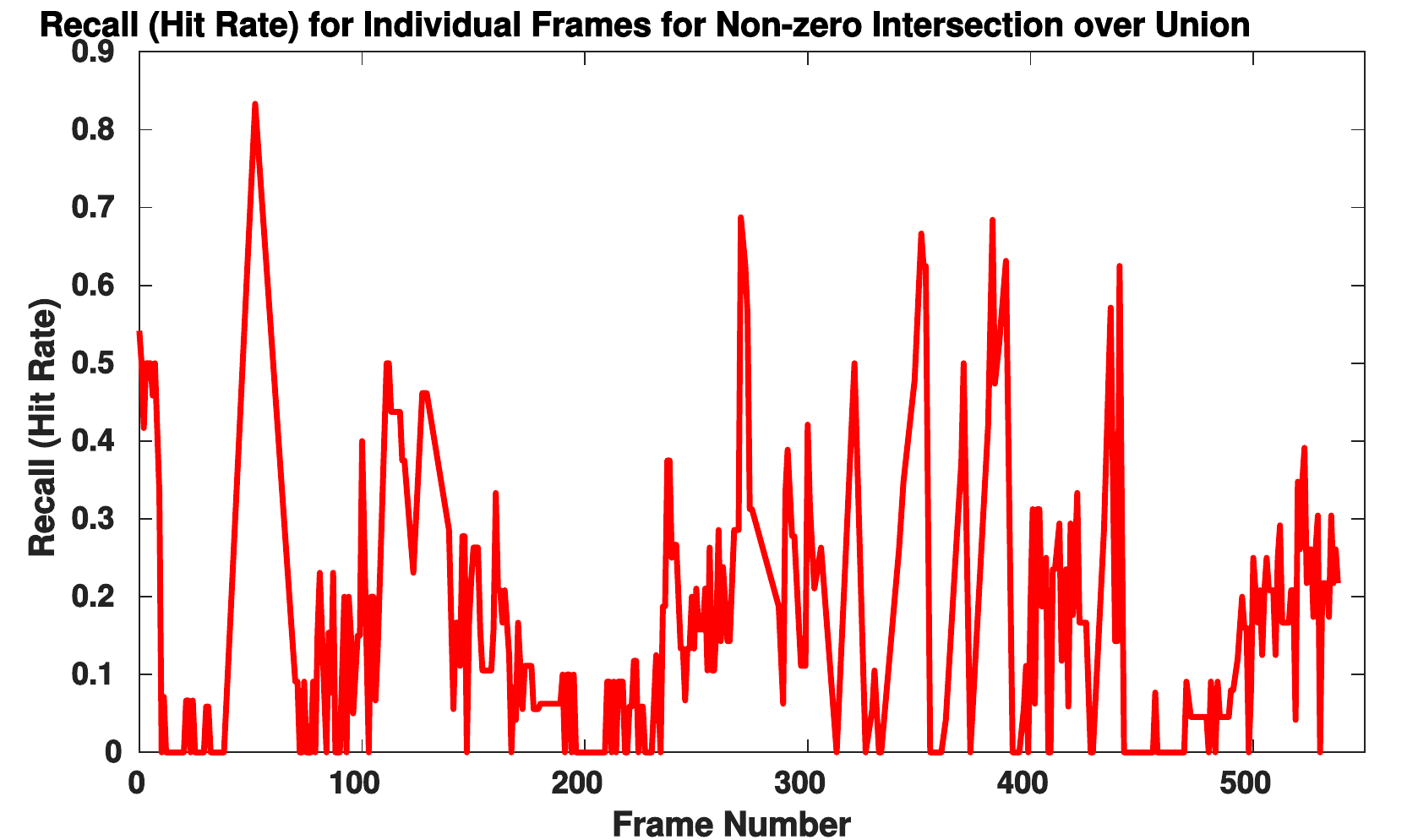}                  
        \caption{Per-frame recall}       
        \label{fig:inherent_fuziness}        
        \end{subfigure}\quad
        \begin{subfigure}[b]{.32 \textwidth}
         \raggedleft
            \includegraphics[width = \textwidth]{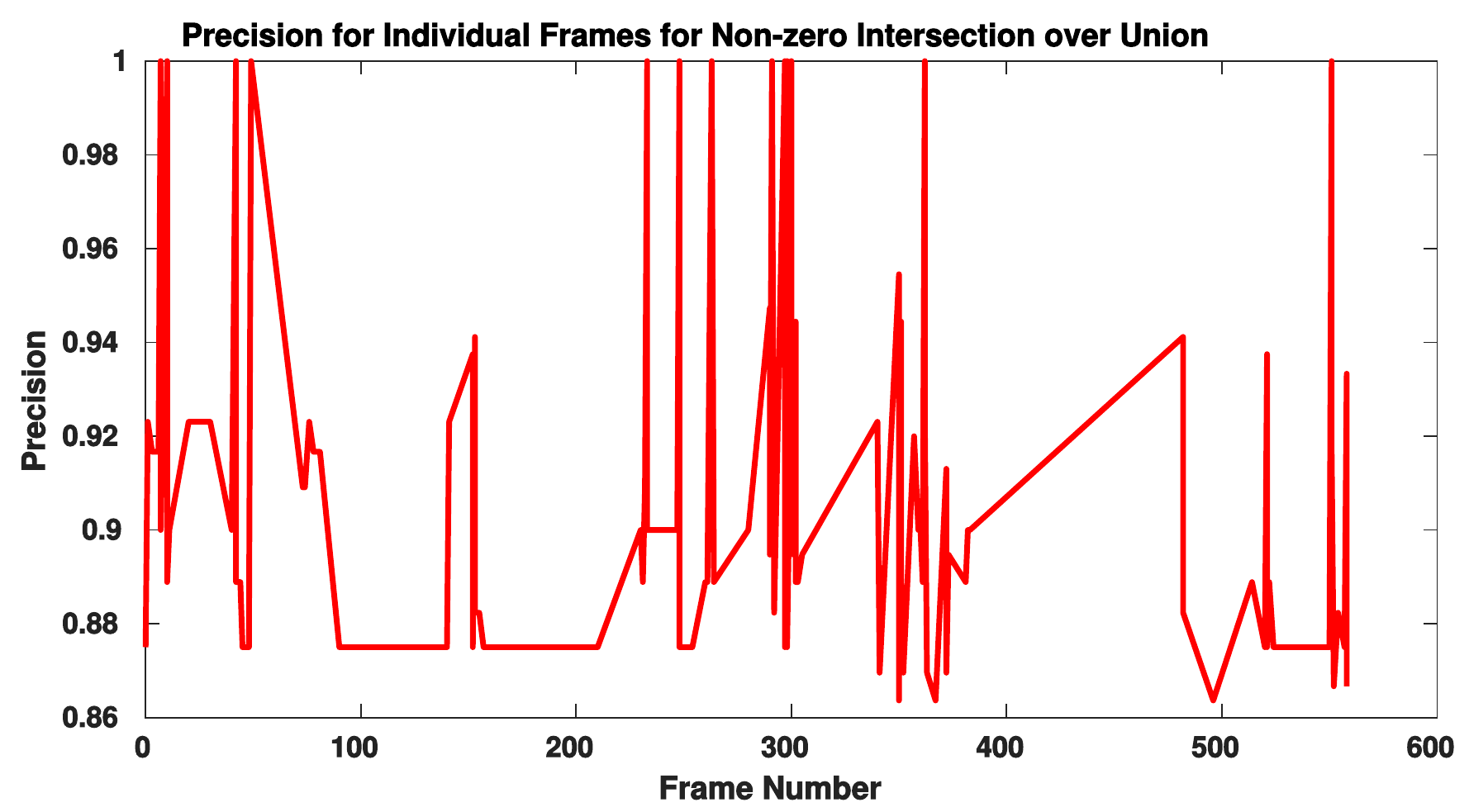}                  
        \caption{Per-frame precision}       
        \label{fig:pframeprecision}        
        \end{subfigure}         
   \caption{Evaluation metric for segmentation, per-frame recall and precision. Figure on left shows how segmentation accuracy is evaluated by performing one to one comparison between the ground truth and our resultant bounding boxes. The figure on the middle, shows the variance in recall for per frame segmentation. The rightmost image shows that, our precision per-frame is always over $87\%$. Coupling this with multiple views we achieve high precision and recall (Fig.~\ref{fig:recgoodmarginal}, \ref{fig:prec}, \ref{fig:rec}, \ref{fig:f1measure}) for the entire videos.}
   \label{fig:eval_metric}
\end{figure*} 
In this section, we study the performance of our segmentation method. In particular, we investigate its sensitivity with respect to user supervision across the validation datasets. We use three metrics, precision, recall and $F_1$- measure for this purpose. 
As is well-known in literature, these metrics are obtained using true positives (TP), false positives (FP) and false negatives (FN). Formally, $\texttt{precision} = \frac {\texttt{TP}}{\texttt{TP}+\texttt{FP}}$ and $\texttt{recall} = \frac {\texttt{TP}}{\texttt{TP}+\texttt{FN}}$  and $F_1$ measure = $2 \left(\frac{\texttt{precision} \cdot \texttt{recall}}{\texttt{precision} + \texttt{recall}} \right)$. We define \texttt{TP, FP} and \texttt{FN} using manually tagged apples and the apples detected by our algorithm. Specifically, \texttt{TP =}  apples detected by our algorithm and tagged manually, \texttt{FP =} apples detected by the algorithm but not tagged manually, \texttt{FN =} apples tagged manually but not detected by our algorithm.


We compare the bounding boxes computed by our algorithm to manually drawn bounding boxes (guidelines for the manual annotation procedure was described in the previous section). To evaluate the quality of a particular bounding box generated by our algorithm, we use a metric well-known in the literature as the intersection over union (IoU) threshold (\cite{intersectionunion}). This method factors in how much of each bounding box is detected by our algorithm (Fig.~\ref{fig:eval_metric}(\subref{fig:segmentation_evaluation})). We demonstrate the performance of our algorithm in terms of precision, recall and $F_1$- measure over the entire range of intersection over union threshold. For the purposes of counting though, we utilize a nominal nonzero intersection over union threshold ($\textbf{0.01}$).

\begin{figure*}[!hbpt]
        \centering     
            \includegraphics[width=\textwidth]{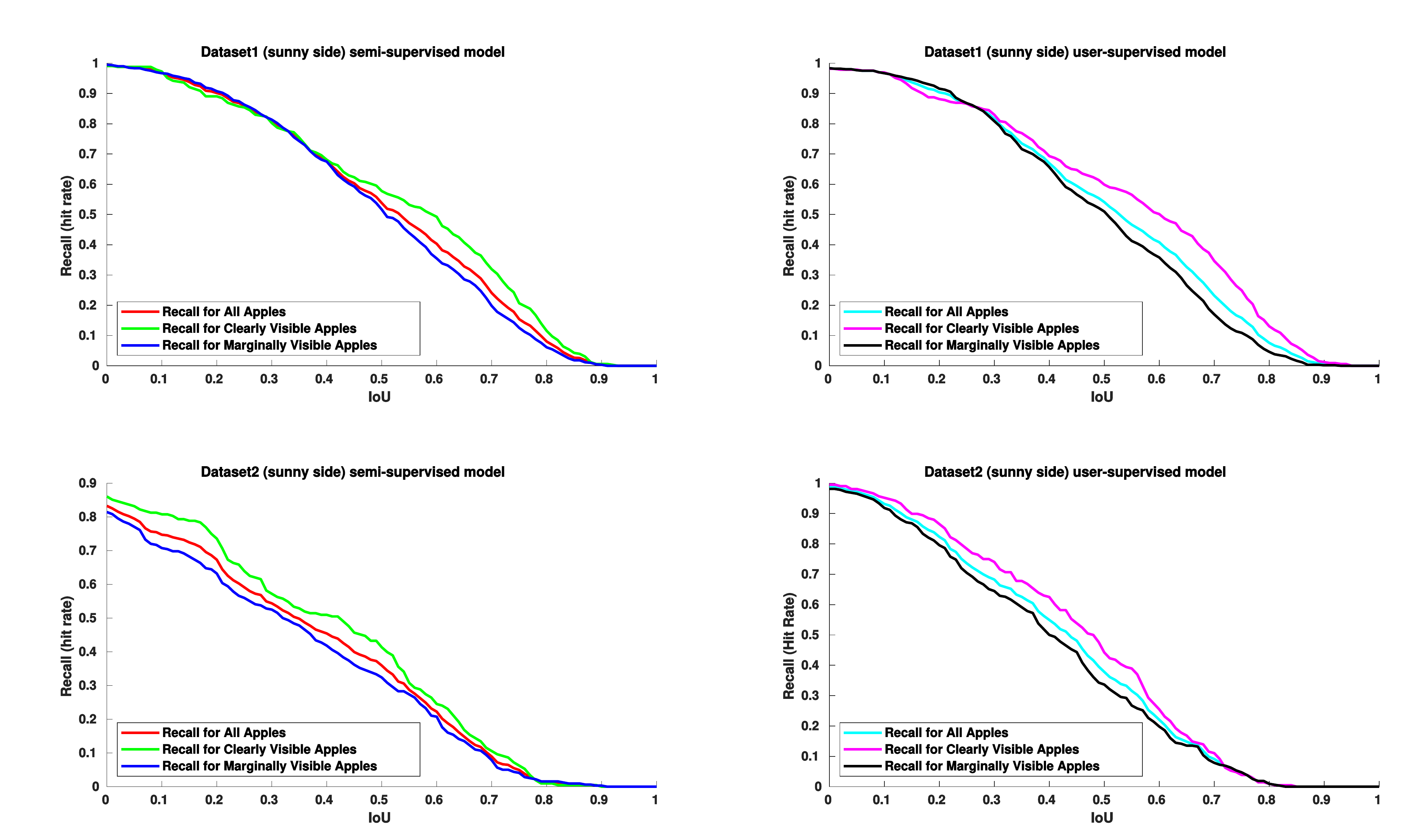}           
            
   \caption{Recall (hit rate) for clearly visible, marginally visible and all apples for both semi-supervised and \supemph{user-supervised} case. As expected, the recall for clearly visible apples are high, and the recall for marginally visible apples are low. The overall recall is in between this low and high bound.}
   \label{fig:recgoodmarginal}
\end{figure*}

\begin{figure*}[!hbpt]
\centering   
    \includegraphics[width=.9\textwidth]{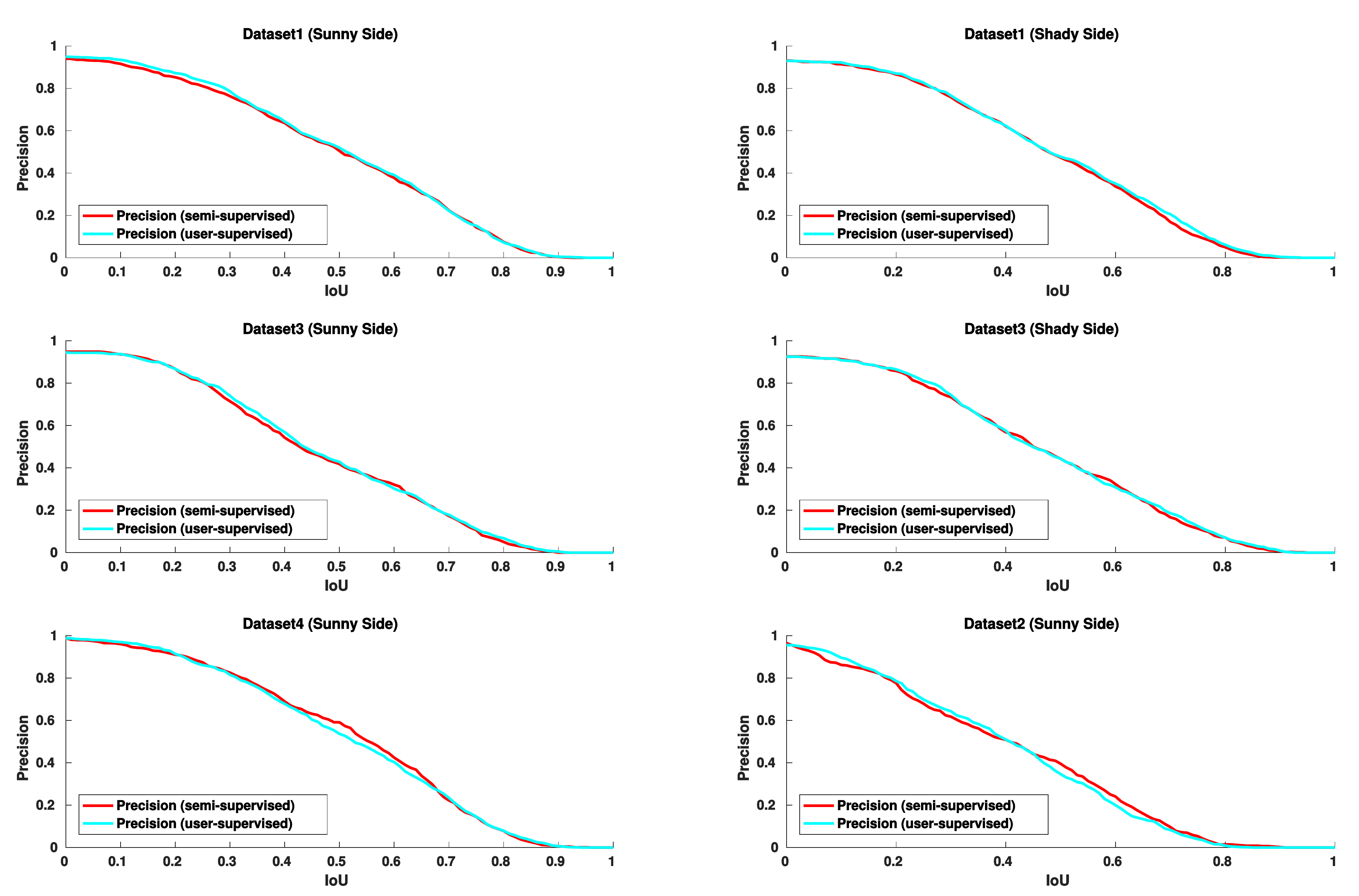}   
     \caption{Precision for \supemph{semi-supervised} and \supemph{user-supervised} models. For all the videos, the obtained precision is similar for both models.}
   \label{fig:prec}
\end{figure*}

\begin{figure*}[!hbpt]
    \centering
    \includegraphics[width=.9\textwidth]{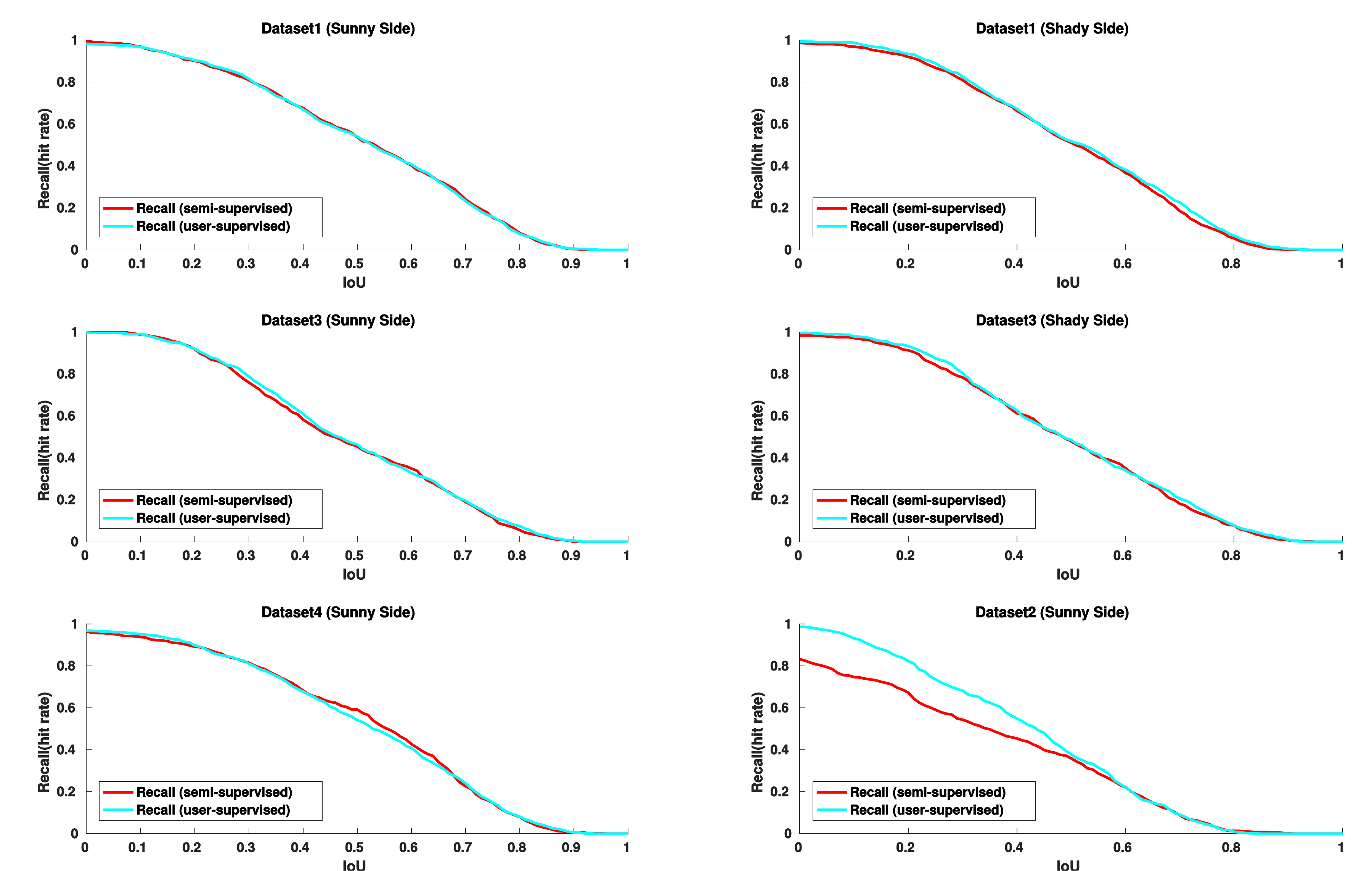}   

   \caption{Recall for \supemph{semi-supervised} and \supemph{user-supervised} models. For the five videos (collected from Dataset1, Dataset3, Dataset4) the recall for both models is similar. For the last video (collected from Dataset2) the recall for the \supemph{semi-supervised} model drops by $20\%$. This is due to the fact that, the apples contained this video are a mixture of green and red which is not captured by the \supemph{semi-supervised} model representing primarily red apples.}
   \label{fig:rec}
\end{figure*}

\begin{figure*}[!hbpt]
        \centering     
            \includegraphics[width=.9\textwidth]{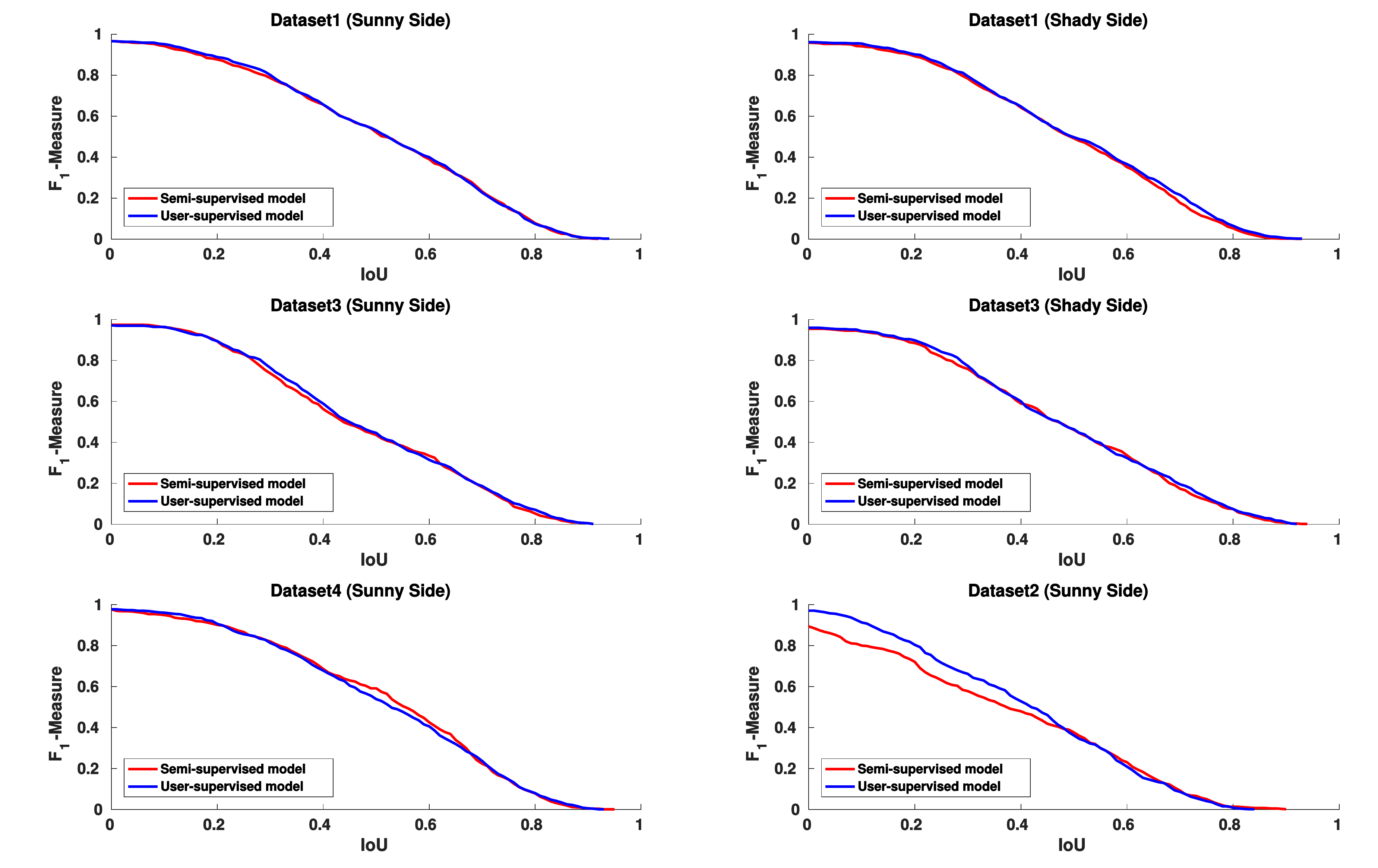}           
            
   \caption{$F_1$ - measure for both \supemph{semi-supervised} and \supemph{user-supervised} models.}
   \label{fig:f1measure}
\end{figure*}

One of the advantages of our algorithm is that, we do not detect all the apples on a single image. We utilize multiple views available from the video. It is evident from Fig.~\ref{fig:eval_metric}(\subref{fig:inherent_fuziness}) that for single frames even for nonzero intersection over union threshold (means if the algorithm detected bounding boxes just touches the ground truth), the per-frame apple detection rate (recall) varies significantly. Nevertheless, our per-frame precision is always over $87\%$ and with the help of multiple views of each apple cluster we achieve high precision and recall ( Fig.~\ref{fig:recgoodmarginal}, \ref{fig:prec}, \ref{fig:rec}, \ref{fig:f1measure}).

In order to test the performance without user intervention, we trained a classification model from the dataset collected in 2015 and used this model for detecting apples from the validation dataset collected in 2016. To test the effect of user supervision on specific videos collected from datasets, we let users choose the apples from the first fifty frames of the input video. We utilize the clusters chosen by the users to build the classification model and use this model for segmenting apples. For the rest of the paper, we will refer to these two types of classification models as \textbf{\emph{semi-supervised}} and \textbf{\emph{user-supervised}} models.


First, we evaluate the computed recall for clearly and marginally visible apples. According to the visibility guidelines discussed in Section~\ref{subsec:manuallabel}, the recall/ hit rate for clearly visible apples should be higher than the recall for marginally visible apples. Fig.~\ref{fig:recgoodmarginal} shows, that this is indeed the case for both the \supemph{semi-supervised} and \supemph{user-supervised} classification models. We show the results for two different videos from the validation datasets (Dataset1 and Dataset2) in Fig.~\ref{fig:recgoodmarginal}. For the figures on the top row, both models achieve high recall. For the figures on bottom row the \supemph{semi-supervised} model has a low recall and the \supemph{user-supervised} model has high recall. Importantly though, for all the figures we see that recall for clearly visible apples are higher than recall for marginally visible apples and the overall recall is in between these two (especially for high IoU thresholds).

Second, we investigate the sensitivity of user input. For the \supemph{user-supervised} case, we only allow choosing apples for the first fifty frames. We see in Fig.~\ref{fig:prec},\ref{fig:rec} for the first five videos collected from Dataset1, Dataset3 and Dataset4 the precision and recall for both the \supemph{user-supervised} and \supemph{semi-supervised} models are similar. However, for the video collected from Dataset2, the apples are a mixture of red and green. Therefore, the \supemph{semi-supervised} classification model does not generalize well. Consequently, the precision is similar but the recall drops by $20\%$.

Finally, it is desired to obtain a single number associated with the performance of the algorithm. We use $F_1$- measure for this purpose. This is the harmonic mean of precision and recall. Our $F_1$-measure ranges from $.95-.97$ for six videos (Fig. \ref{fig:f1measure}, Table~\ref{tab:f1}) for nonzero intersection over union threshold. The best known $F_1$-measure $.91$ was obtained by (\cite{deepApple}). As the datasets are different, and our method does not maximize detection for a single image, a direct comparison is not possible. In the next section, we investigate the performance of our counting algorithm.  
\begin{table}
\begin{tabular}{|c|c|c|}
\hline 
 Datasets & \supemph{Semi-supervised} & \supemph{User-supervised} \\ 
\hline 
 $D_1$ (Sunny) & .9662 & .9658 \\ 
\hline 
 $D_1$ (Shady) & .9585 & .9609 \\ 
\hline 
 $D_3$ (Sunny) & .9738 & .9711 \\ 
\hline 
$D_3$ (Shady) & .9541 & .9592 \\ 
\hline 
$D_4$ (Sunny) & .9774 & .9775 \\ 
\hline 
$D_2$ (Sunny) & .8931 & .9710 \\ 
\hline 
\end{tabular} 
\caption{$F_1$- Measure for the Validation Dataset for IoU = .01. Here, $D_1, D_2, D_3, D_4$ denote Dataset1, Dataset2, Dataset3 and Dataset4.}
\label{tab:f1}
\end{table}

\subsection{Performance Evaluation of the Counting Method}\label{subsec:count_res}
In this section, we quantify the performance of our counting method. We investigate how the algorithm (per-frame counting) performs on individual segmented images and how the overall performance (merging counts across frames) compares to human-perceived counts and ground truth. It is notable that following the segmentation step, all the bounding boxes with above nonzero intersection over union threshold are used in the counting step.

For evaluating both per frame counting and merging, we utilize the videos collected from the validation datasets (Dataset1 - Dataset4). We start with the per-frame counting method.

\subsubsection{Evaluation of Per-frame Counting Method:} To evaluate the performance of the per-frame counting algorithm, we took all the segmented images from seven videos collected from the four datasets. Afterward, we performed a connected component analysis on them, randomly selected $5000$ components and marked each one with the perceived count from a human point of view. These counts are then compared to the counts obtained from the algorithm. At this stage, we want the segmented images to be accurate and consequently, we use the \supemph{user-supervised} model for segmentation.
\begin{figure*}[!hbpt]
        \centering
        \begin{subfigure}[b]{\textwidth}\begin{subfigure}[b]{.55\textwidth}
                 \includegraphics[width=\textwidth]{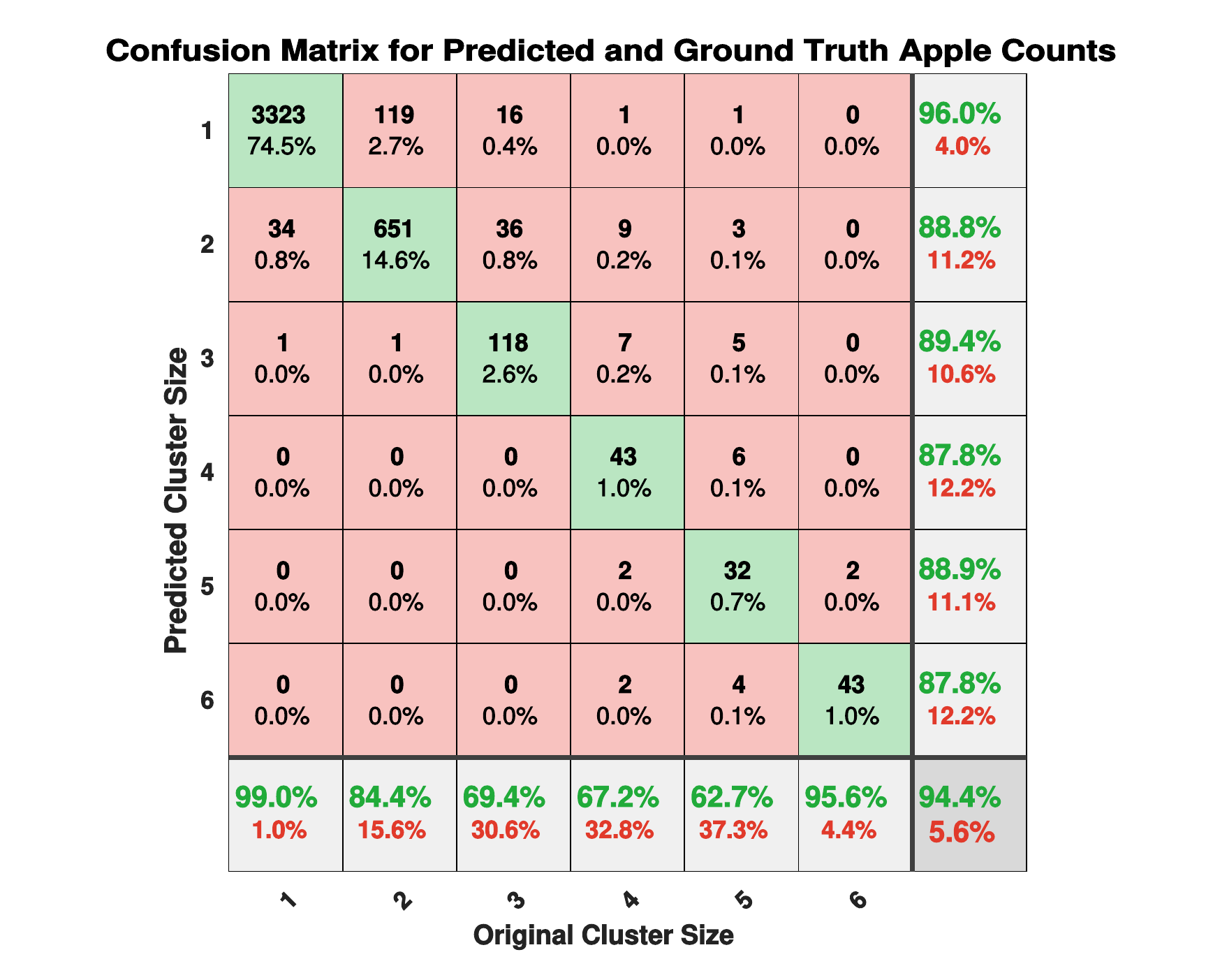}
                 \caption{Confusion matrix for the per-frame counting method.}
                     \label{fig:confusionmat}   
        \end{subfigure}\quad \begin{subfigure}[b]{.40\textwidth}
                         \includegraphics[width=\textwidth]{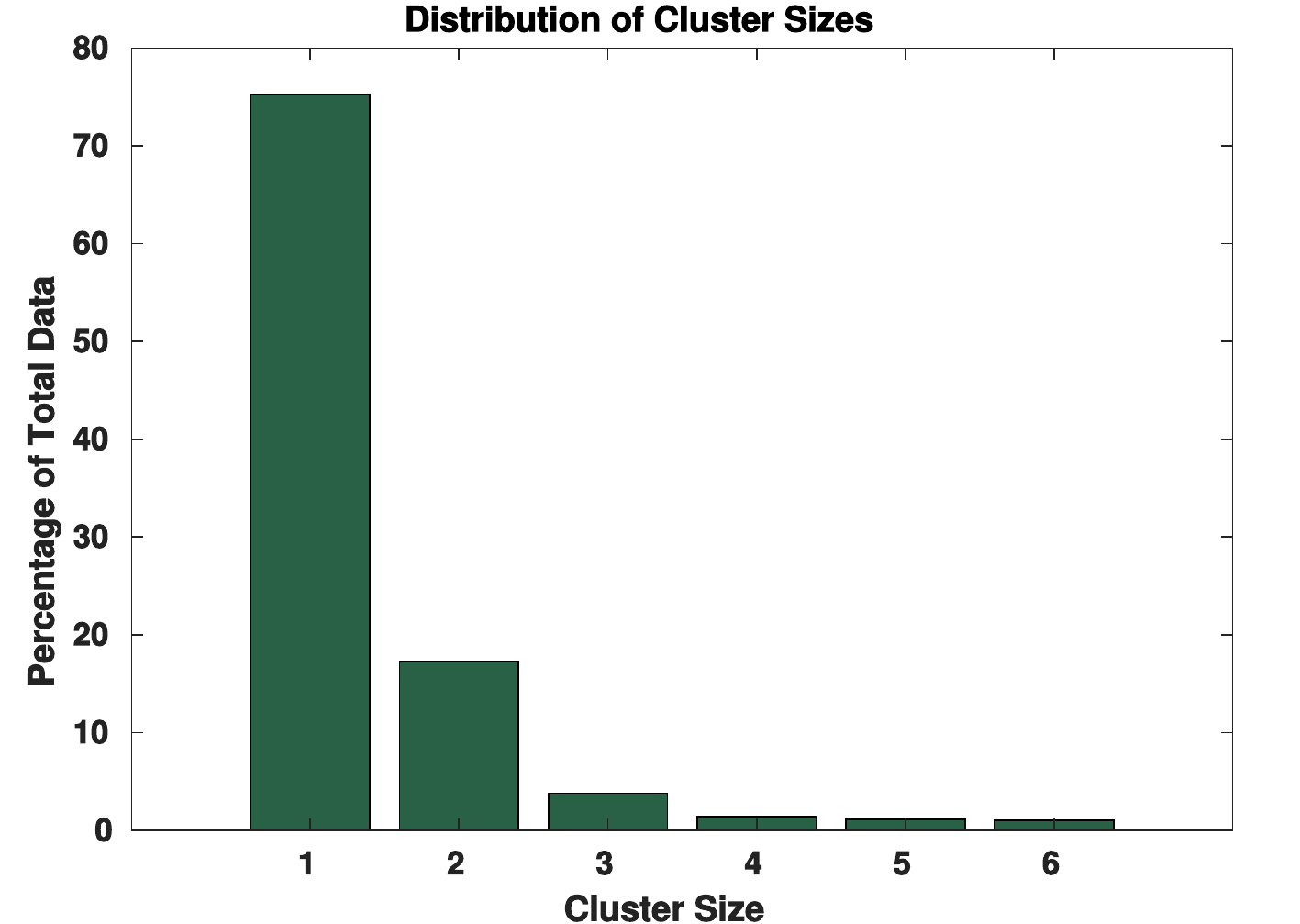}
                 \caption{Distribution of cluster sizes.}
                     \label{fig:clusterDist}   
        \end{subfigure}
        \end{subfigure}\\ \begin{subfigure}[b]{\textwidth}
        \includegraphics[width=\textwidth]{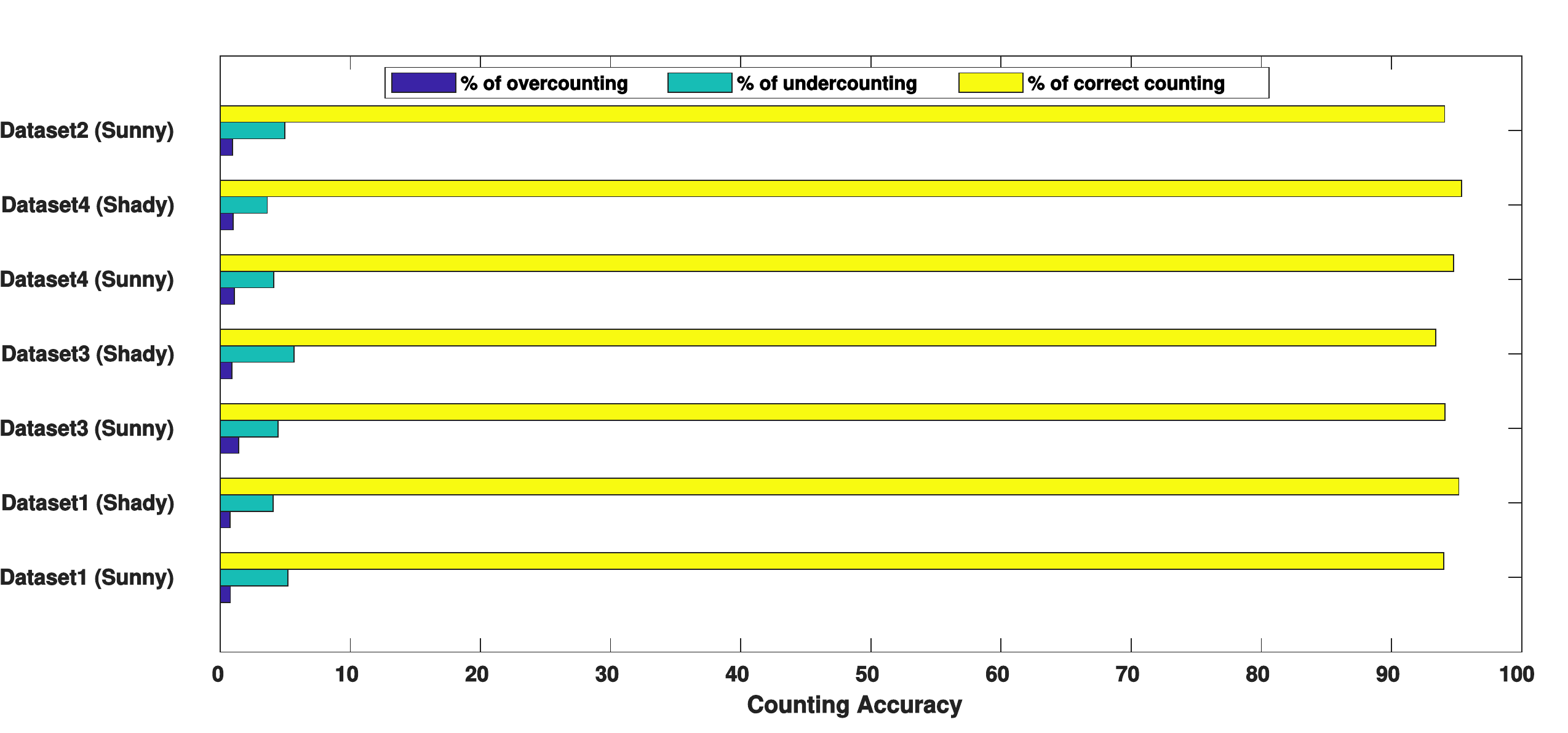} 
        \caption{Accuracy of the per-frame counting method across different videos collected from Dataset1-Dataset4.}
             \label{fig:countdatasets}
        \end{subfigure}
   \caption{Evaluating the performance of the per-frame counting algorithm- confusion matrix (left), distribution of cluster sizes (middle) and performance across different datasets (right). The right-bottom cell in the confusion matrix (figure on left) shows that the overall accuracy of our method is $94.4\%$. The rightmost column (Rows $1-6$) shows the precision per cluster size and the bottom row (Columns $1-6$) shows recall per cluster size. From the figure on the middle, we find that single apple clusters are dominating the data ($75.31\%$). The figure on right shows how our counting accuracy varied from $94.01\% - 95.38\%$ across different videos.}
   \label{fig:analysis}
\end{figure*}    
Essentially, we have three key insights from this experiment. First, it is evident from the confusion matrix (Fig.~\ref{fig:analysis}(\subref{fig:confusionmat})), that recall drops with increasing cluster size (varies from $62.7\% - 99\%$) but precision stays over $87\%$ for any cluster size (varies from $87.8\% - 96\%$). Second, for a large portion of the data - single apples ($75.31\%$ of entire data (Fig.~\ref{fig:analysis}(\subref{fig:clusterDist}))); the precision and recall of our algorithm are $96\%$ and $99\%$ respectively. Consequently, the overall accuracy of our method is high ($94.4\%$) (shown in the right-bottom cell in the confusion matrix). Third, low recall rates for larger clusters does not affect the overall performance. In the next section, utilizing the high precision and multiple views we achieve good recall on the entire data (Fig.~\ref{fig:countingtrack}(\subref{fig:dsetcountmodel})).

Next, we quantify the effect of lighting conditions (sunny, shady, cloudy etc.) on the counting method. We computed the accuracy of the per-frame counting method across all the collected videos (which were collected in different lighting conditions). Our counting accuracy varied from $94.01\% - 95.38\%$. Undercounting percentage varies from $3.62\%- 5.67\%$ and overcounting varies from $.74\% - 1.1\%$. These results are presented in Fig.~\ref{fig:analysis}(\subref{fig:countdatasets}). In the next section, we evaluate the performance of merging the apple counts using camera motion.

\subsubsection{Merging Counts Across Multiple Frames}To verify the performance of the merging method, we utilize the manual annotations. We treat the manually annotated fruits as human perceived ground truth. Afterward, we track these fruits across frames using camera motion (3D camera poses) to avoid double counting and find the number of unique apples. The counts obtained in this manner are then compared to the counts from our merging algorithm.
\begin{figure*}[!hbpt]
        \centering
        \begin{subfigure}[b]{\textwidth}\begin{subfigure}[b]{.49\textwidth}
                 \includegraphics[width=\textwidth]{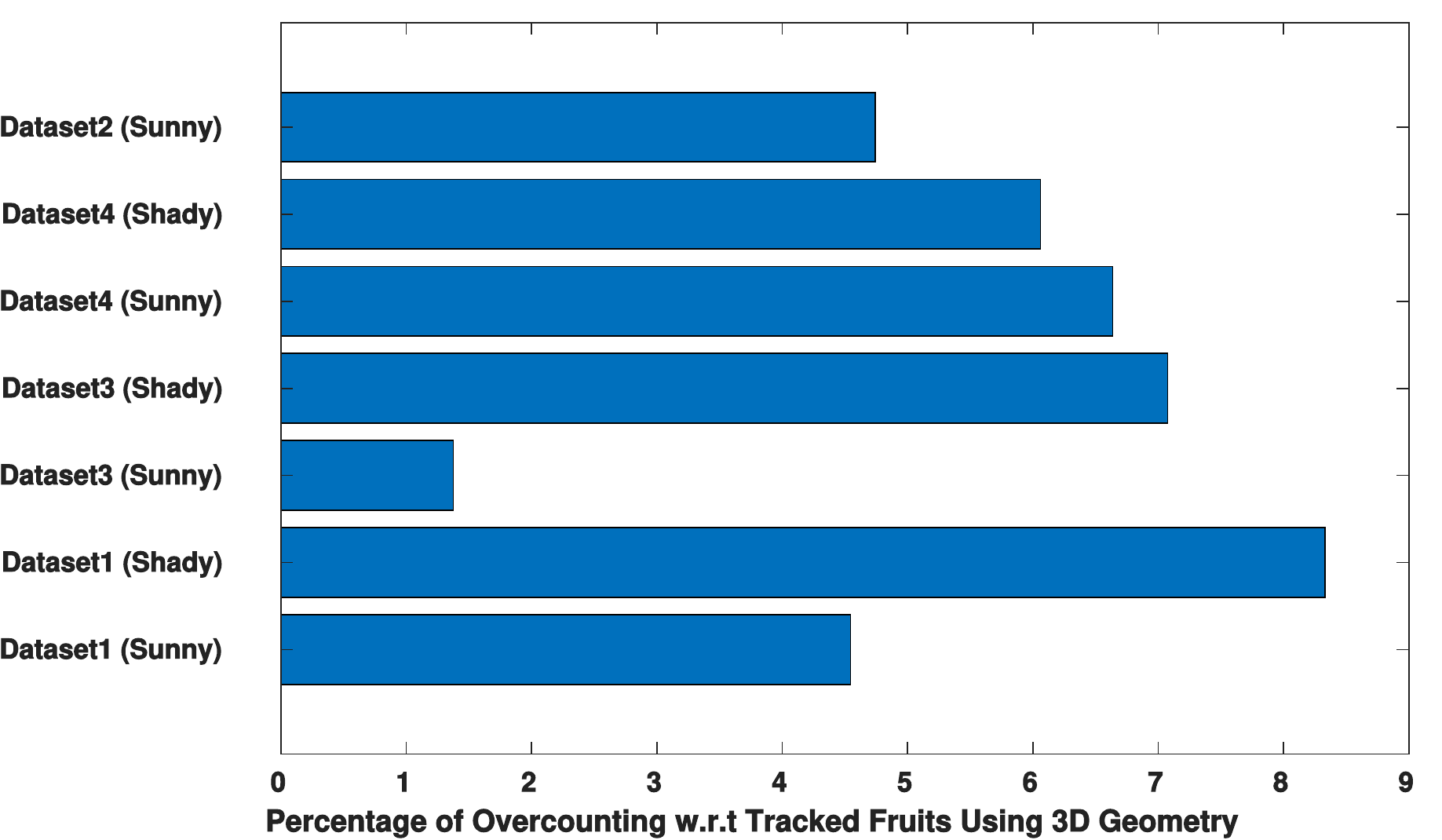}
                 \caption{Evaluation of tracking by homography.}
                     \label{fig:trackHom}   
        \end{subfigure}\quad \begin{subfigure}[b]{.49\textwidth}
                         \includegraphics[width=\textwidth]{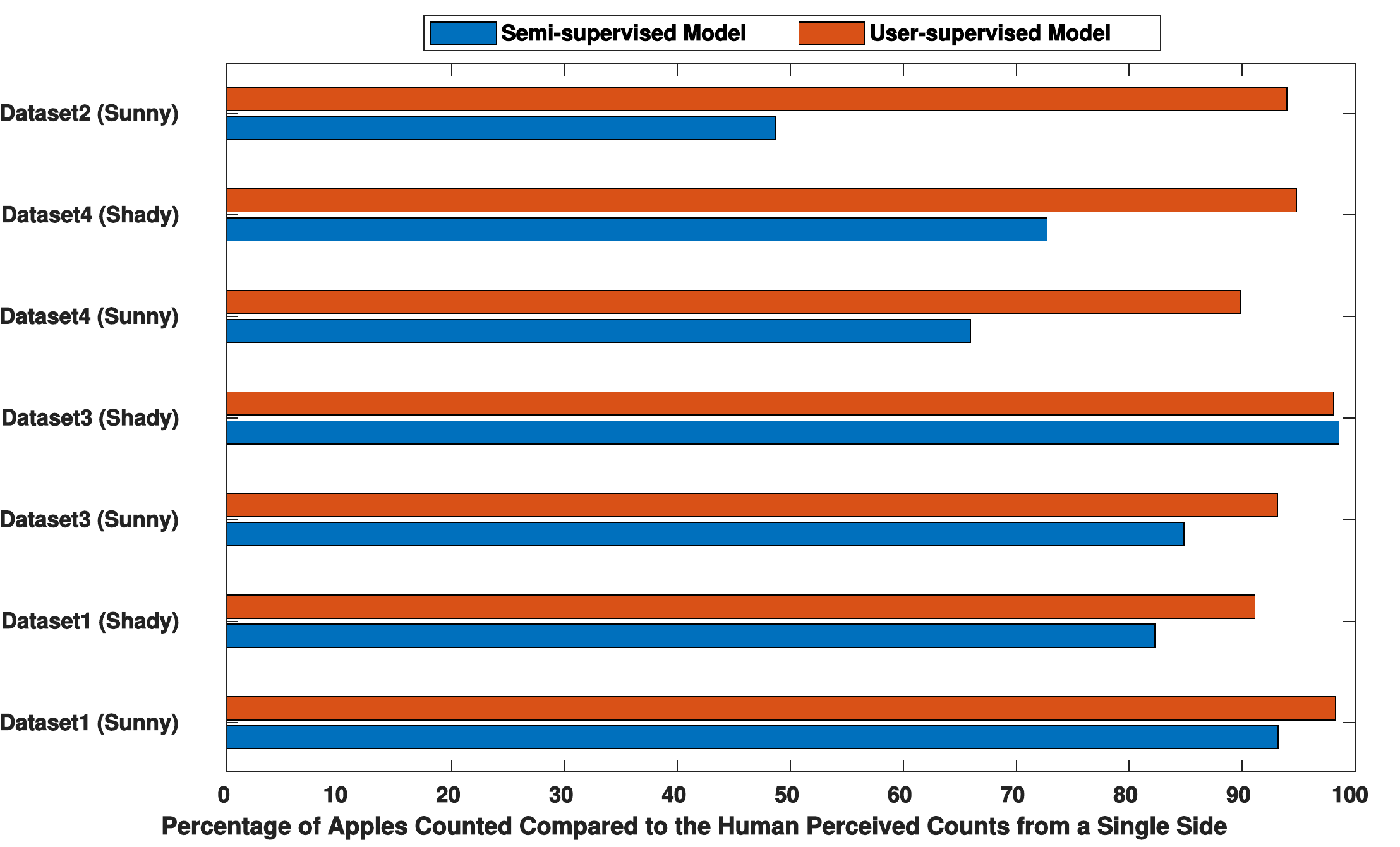}
                 \caption{Performance comparison of \supemph{semi-supervised} and \supemph{user-supervised} models for counting.}
                     \label{fig:dsetcountmodel}   
        \end{subfigure}
        \end{subfigure}       
   \caption{Evaluating the performance of the counting algorithm along with tracking. The figure on left shows the percentage of overcounting due to using homography for tracking (up to $8\%$). Figure on the right shows, the percentage of visible apples counted by our method using both \supemph{semi-supervised} and \supemph{user-supervised} models.}
   \label{fig:countingtrack}
\end{figure*}
First, we quantify the amount of overcounting due to approximating the camera motion using homography. We performed this evaluation by simply comparing the number of unique manually labeled apples obtained from homography to the number of unique apples obtained by utilizing full 3D camera motion. The camera motions were computed using a commercial photogrammetry software Agisoft (\cite{agisoft}). We found that across different datasets; using homography can lead to $8\%$ overcounting. Fig.~\ref{fig:countingtrack}(\subref{fig:trackHom}) shows these results. 

Next, we evaluate the accuracy of the merged counts (tracking by homography) to human-perceived counts (tracking by 3D camera poses). To understand the importance of user interaction, we perform this analysis for both \supemph{semi-supervised} and \supemph{user-supervised} models (Fig.~\ref{fig:countingtrack}(\subref{fig:dsetcountmodel})). Our accuracy with respect to human perceived ground truth varies from $89\% - 98\%$ for the \supemph{user-supervised} model and $48\% - 98\%$ for the \supemph{semi-supervised} model. The drop in accuracy for the semi-supervised model was propagated from the segmentation phase. The main takeaways from these results are, 1)~accurate segmentation is very important for obtaining correct counts. 2)~based on the geometry of the environment and lighting conditions, we count $89\% - 98\%$ of the visible apples from a single side. In the next section, we investigate how the single side counts correlate with the actual yield.

\subsection{Yield Estimation}
\begin{figure*}[!hbpt]
        \centering
                \begin{subfigure}[b]{\textwidth}\begin{subfigure}[b]{.485\textwidth}
                 \includegraphics[width=\textwidth]{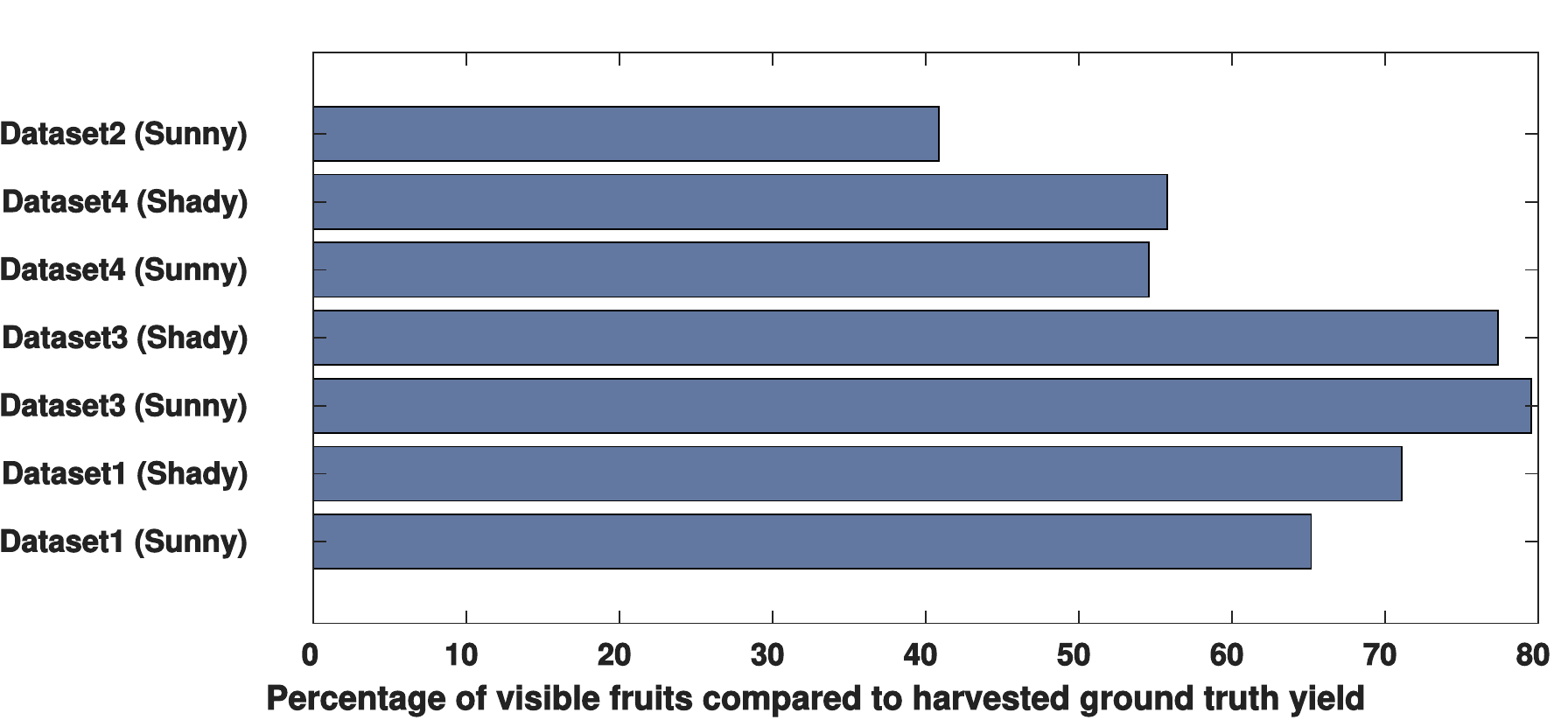}
                 \caption{Percentage of visible apples from a single side compared to ground truth yield.}
                     \label{fig:labelgt}   
        \end{subfigure}\quad \begin{subfigure}[b]{.485\textwidth}
                         \includegraphics[width=\textwidth]{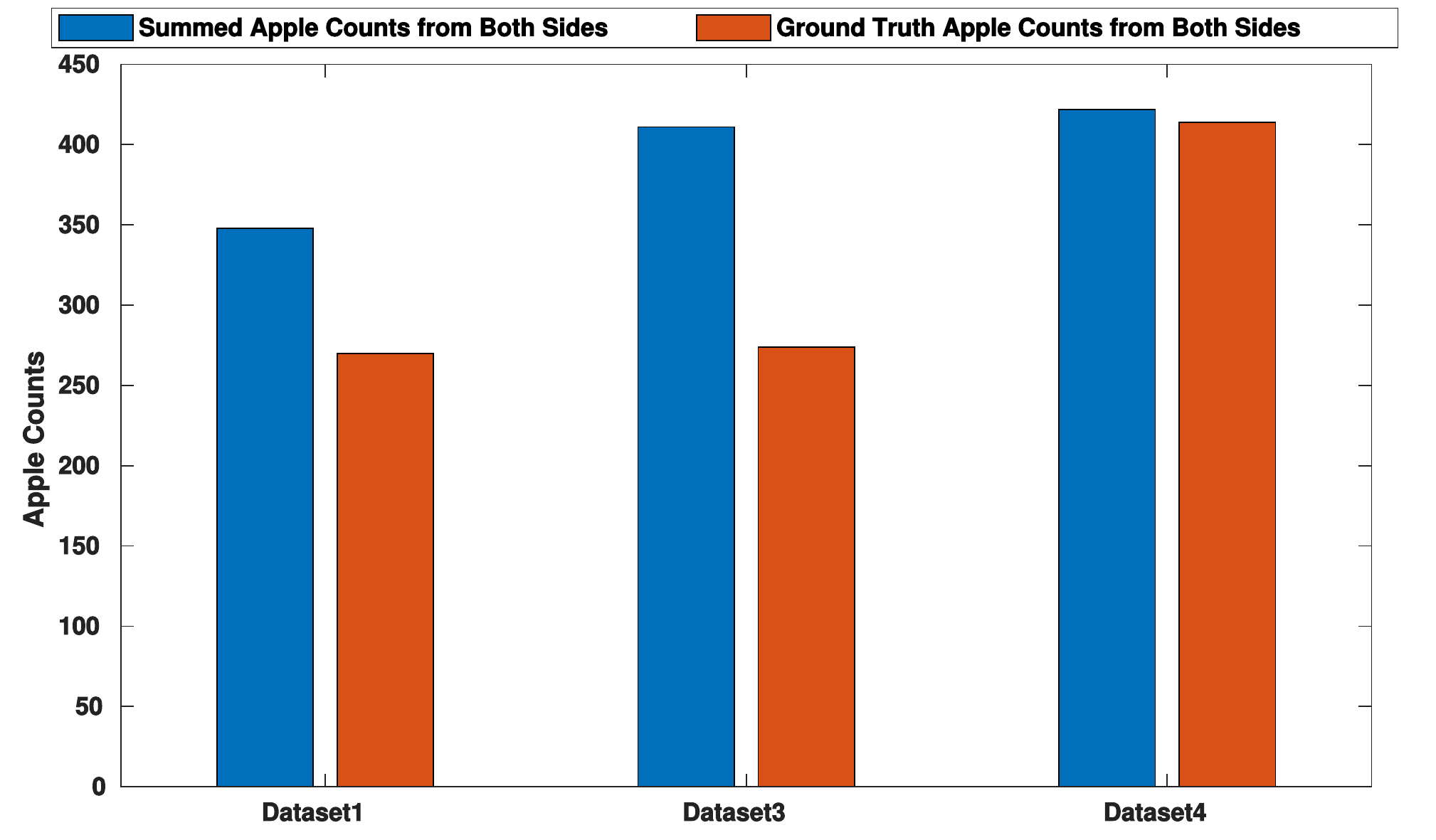}
                 \caption{Summed counts from single sides compared to obtained ground truth.}
                     \label{fig:yieldsum}   
        \end{subfigure}
        \end{subfigure}
   \caption{Correlation between single side fruit counts and actual yield. The figure on left shows the percentage of visible apples from a single side compared to ground truth ($40.85\% - 79.83\%$). Figure on the right shows the summed fruit counts from both sides compared to ground truth fruit counts ($101.93\% - 150\%$).}
   \label{fig:singlesideyield}
\end{figure*}

Our original goal is to get an accurate yield estimate. For that, we need to correlate the counts from our algorithm (from the single side of a row) to the actual ground truth (number of harvested fruits). Toward this, we first try to find out a correlation between the number of visible apples from a single side and the actual yield. As mentioned earlier, we determine the total number of visible apples by tracking the manually labeled apples from a single side, utilizing estimated camera motion. For this step, we will use datasets where videos from both sides were collected (Dataset1, Dataset3, Dataset4). From Fig.~\ref{fig:singlesideyield}(\subref{fig:labelgt}), it is evident that the number of visible apples from a single side vary greatly across different datasets ($40.85\% - 79.83\%$). This is expected as the orchard from which we collected the data is not well trimmed and the size and shape of the trees varies significantly even within the same row. 
\begin{figure}[!hbpt]
        \centering
            \includegraphics[width =0.85\columnwidth]{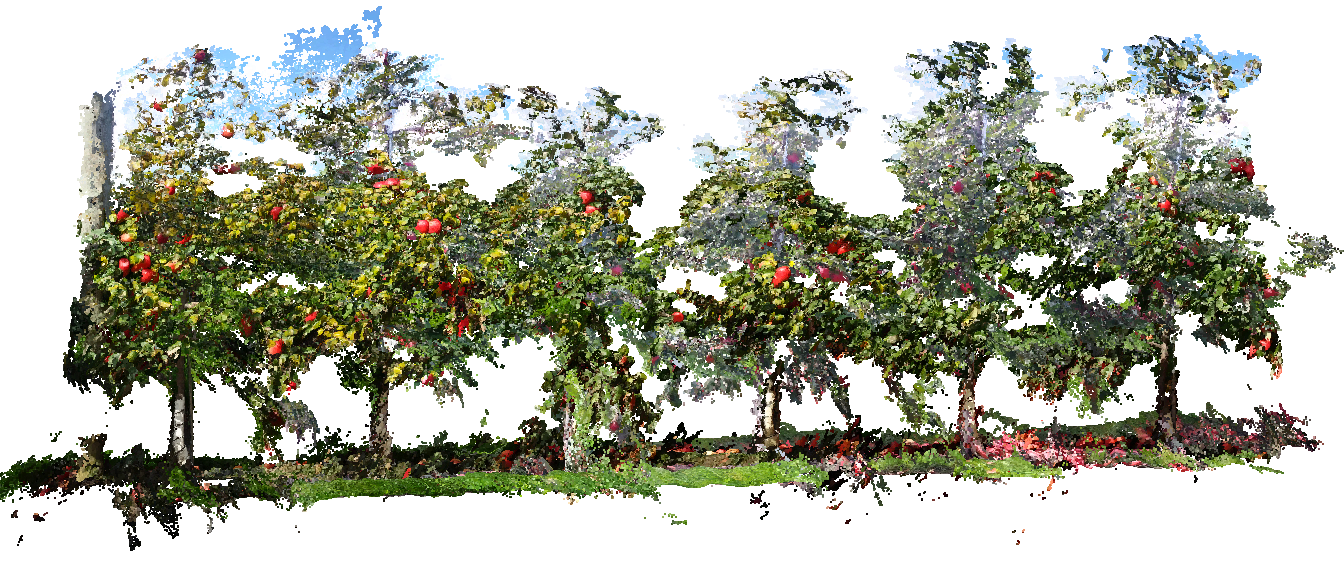}           
   \caption{Merged reconstruction for Dataset1.}
   \label{fig:mergeboth}
\end{figure}

Another simple solution is adding the apple counts from both sides and finding a correlation with the actual yield. Fig.~\ref{fig:singlesideyield}(\subref{fig:yieldsum}) shows these results and again, the summed yields vary considerably across datasets ($101.93\% - 150\%$). Therefore, to get a useful estimate we need to merge the fruit counts from both sides of the tree rows. Next, we discuss this procedure in details.

To merge counts from both sides, we utilize the 3D geometry of the environment. We reconstruct each side of the row using captured images. We use a photogrammetry software Agisoft (\cite{agisoft}) for this purpose. Afterward, we merge the reconstructions from both sides using semantic constraints (\cite{dong2018treeSBA,techreportroy}). Fig.~\ref{fig:mergeboth} shows an example (merged reconstruction for Dataset1).

\begin{figure*}[!hbpt]
        \centering
            \includegraphics[width=0.90\textwidth]{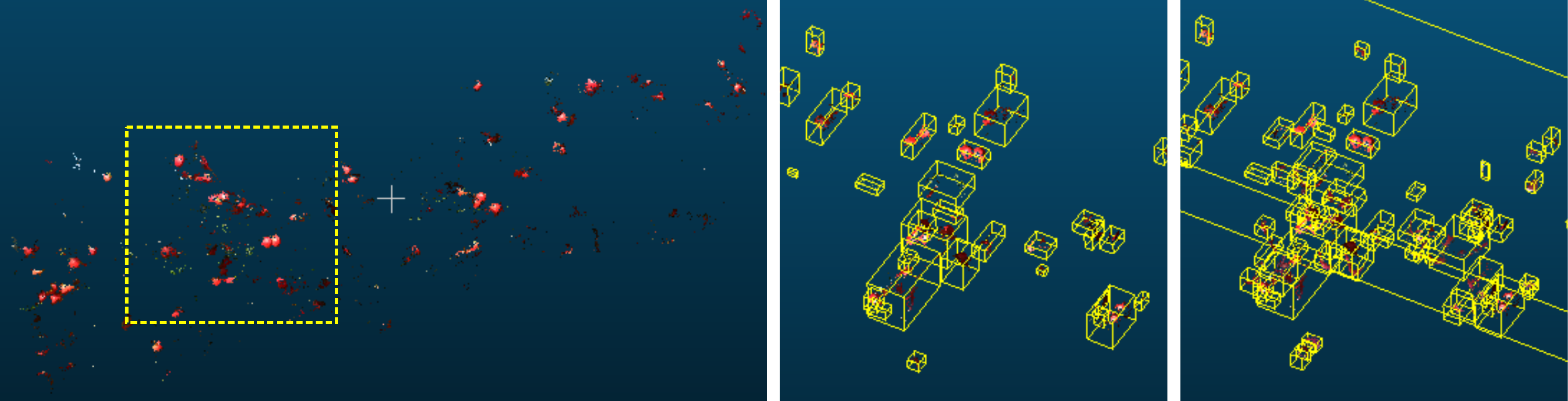}           
   \caption{Apples in 3D and intersecting clusters from both sides. Left figure shows the apples from the front side of Dataset1 detected in 3D. Figure in the middle shows the computed connected components on this apples. Figure in the right shows how the connected components from the back are intersecting with the front ones.}
   \label{fig:applecountingbothsides}
\end{figure*}

Next, we detect the fruits using our segmentation method (Section~\ref{sec:segmentation}) and back-project the detected fruits in the images to obtain the fruit location in the 3D reconstruction. Fig.~\ref{fig:applecountingbothsides} shows an example. 
We perform a connected component analysis to detect the apple clusters in 3D. Then we project individual 3D clusters back to the images by utilizing the recovered camera motion. We count the fruits from these reprojected images using our counting method developed in Section~\ref{sec:counting}. A 3D cluster can be tracked over many frames. We choose three frames with the highest amount of detected apple pixels (from the 3D cluster) and report the median count of these three frames as the fruit count for the cluster. We follow this procedure for all the detected 3D clusters and aggregate the fruit count from a single side. 

\begin{figure}[!hbpt]
        \centering
            \includegraphics[width =0.90\columnwidth]{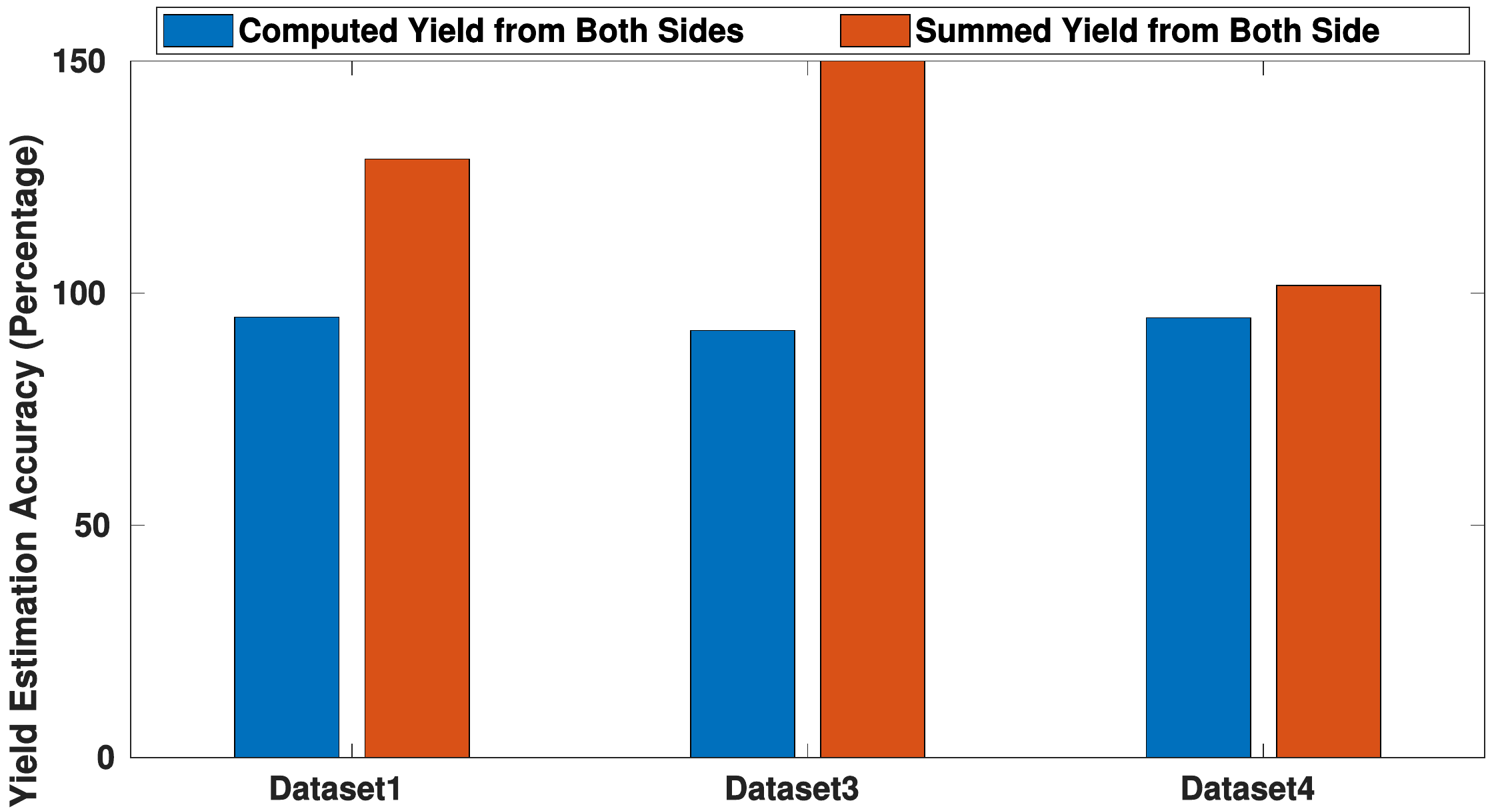}           
   \caption{Total fruit counts computed from the merged reconstruction and independent single side reconstructions compared to ground truth. The single side fruit counts are computed by summing up independent single side counts ($101.93\%- 150\%$). Evidently, the counts from merged reconstructions are more consistent ($91.98\% - 94.81\%$).}
   \label{fig:fruitCount}
\end{figure}

To merge the counts from both sides, we compute the intersection of the connected components from both sides. Afterward, we compute the total counts by using the inclusion-exclusion principle (\cite{andreescu2004inclusion}). Essentially, we sum up the counts from all the connected components, compute the intersections area among them and add/subtract the weighted parts accordingly. Fig.~\ref{fig:fruitCount} shows our result. Our counting accuracy from both sides for Dataset1, Dataset3 and Dataset4 are $94.81\%,91.98\%,\text{and } 94.68\%$ respectively. Compared to both side counts, if we just add the single side counts we overcount significantly ($128.89\%,150\%,\text{and } 101.93\%$ for Dataset1, Dataset3, and Dataset4 respectively). Table~\ref{tab:yield} summarizes the final yield result. It indicates that merging the rows from both sides is essential to obtain accurate yield.

\begin{table*}
\centering
\begin{tabular}{|c|c|c|c|}
\hline 
\rule[-1ex]{0pt}{2.5ex}  & Harvested fruit counts & Merged fruit counts from both sides & Sum of fruit counts from single sides \\ 
\hline 
\rule[-1ex]{0pt}{2.5ex} Dataset1 & $270$ & $256$ ($94.81\%$) & $348$ ($128.89\%$) \\ 
\hline 
\rule[-1ex]{0pt}{2.5ex} Dataset3 & $274$ & $252$ ($91.98\%$) & $411$ ($150\%$)\\ 
\hline 
\rule[-1ex]{0pt}{2.5ex} Dataset4 & $414$ & $392$ ($94.68\%$) & $422$ ($101.93\%$)\\ 
\hline 
\end{tabular} 
\caption{Summary of yield results.}
\label{tab:yield}
\end{table*}

\section{Conclusion and Future Work}\label{sec:conc}
In this paper, we presented a complete yield estimation system for apple orchards from monocular images. From a purely technical point of view, our main contributions are a semi-supervised clustering method relying on color for identifying the apples and an unsupervised clustering method relying on shape to estimate the number of apples in a cluster. We verified the performance of our algorithms on multiple small datasets. Results indicate that these algorithms perform well in practice and outperform most of the existing methods in terms of detection and counting accuracy. 

As reported in section~\ref{subsec:count_res}, we count $89\% - 98\%$ of the visible apples from a single side of the row in different datasets. To be of practical usage though, we needed to correlate this single side counts with harvested yield. With the help of our recent work (\cite{dong2018treeSBA,techreportroy}), we merged the fruit counts from both sides of fruit tree rows. Our method achieved a varying accuracy of $91.98\% - 94.81\%$ across different datasets. 

In future, we would like to couple our system more closely with the 3D geometry of the environment. We would like to develop techniques to localize each individual apple in a cluster, find the pose of the fruit and measure its diameter.

\section*{Acknowledgement}
The authors thank Joshua Anderson, Professors Emily Hoover, and Cindy Tong from the Department of Horticultural Science, University of Minnesota, for their expertise and help with the experiments. This work is supported in part by NSF grant \# 1317788, USDA NIFA MIN-98-G02 and the MnDrive initiative.
\section*{References}

\bibliography{paperrefs}

\end{document}